\documentclass{article}

\usepackage[preprint]{neurips_2024}
\usepackage[utf8]{inputenc} % allow utf-8 input
\usepackage[T1]{fontenc}    % use 8-bit T1 fonts
\usepackage{hyperref}       % hyperlinks
\usepackage{url}            % simple URL typesetting
\usepackage{booktabs}       % professional-quality tables
\usepackage{nicefrac}       % compact symbols for 1/2, etc.
\usepackage{microtype}      % microtypography
\usepackage{xcolor}         % colors

\usepackage{graphicx}
\usepackage{color}
\usepackage{wrapfig}
\usepackage{enumitem}

\newcommand{\trans}{^\dagger}
\newcommand{\inverse}{^{-1}}
\newcommand{\trace}{\operatorname {Tr}}
\newcommand{\RC}{\mathbb{C}}

\usepackage{amsmath,amssymb,amsfonts,amsthm}

\theoremstyle{plain}
\newtheorem{theorem}{Theorem}[section]
\newtheorem{proposition}[theorem]{Proposition}
\newtheorem{lemma}[theorem]{Lemma}

\newtheorem{conjecture}[theorem]{Conjecture}
\newtheorem{observation}[theorem]{Observation}
\theoremstyle{definition}
\newtheorem{definition}[theorem]{Definition}

\theoremstyle{remark}

\definecolor{white}{rgb}{1,1,1}
\definecolor{blue}{rgb}{0.1,0.1,1}
\definecolor{cyan}{rgb}{0.55,0.6,0.8}  %{0.0,0.7,1}
\definecolor{red}{rgb}{0.8,0.2,0.2}  %{1,0.1,0.1}
\definecolor{green}{rgb}{0,0.8,0} %{0.2,0.6,0.2} 
\definecolor{purple}{rgb}{0.6,0,1}
\definecolor{orange}{rgb}{1,0.6,0.3}

\newcommand{\blue}{\color{black}}
\newcommand{\cyan}{\color{black}}
\newcommand{\red}{\color{black}} 
\newcommand{\green}{\color{black}}

\title{  Discovering Abstract Symbolic Relations by \\ Learning Unitary Group Representations}

\author{%
  Dongsung Huh\\ %\thanks{(webpage, alternative address)} \\
  MIT-IBM Watson AI Lab, Cambridge MA, USA 02139 \\
  \texttt{huh@ibm.com} \\
}

\begin{document}

\maketitle

\begin{abstract}
We investigate a principled approach for symbolic operation completion (SOC), a minimal task for studying symbolic reasoning. While conceptually similar to matrix completion, SOC poses a unique challenge in modeling abstract relationships between discrete symbols. We demonstrate that SOC can be efficiently solved by a minimal model --- a bilinear map --- with a novel factorized architecture. Inspired by group representation theory, this architecture leverages matrix embeddings of symbols, modeling each symbol as an operator that dynamically influences others. 
Our model achieves perfect test accuracy on SOC with comparable or superior sample efficiency to Transformer baselines across most datasets, while boasting significantly faster learning speeds (100{\raise.25ex\hbox{$\scriptstyle\sim$}}1000$\times$). Crucially, the model exhibits an implicit bias towards learning general group structures, precisely discovering the unitary representations of underlying groups. This remarkable property not only confers interpretability but also significant implications for automatic symmetry discovery in geometric deep learning.
% discovering the core structure of symmetry-aware equivariant neural networks.
%
Overall, our work establishes group theory as a powerful guiding principle for discovering abstract algebraic structures in deep learning, and showcases matrix representations as a compelling alternative to traditional vector embeddings for modeling symbolic relationships. %dynamic relationships between symbols.

\end{abstract}
\vspace{-0.4cm} 
% 
% {\cyan In contrast to the low-rank bias used in matrix completion, we demonstrate that SOC can be effectively solved by promoting full-rank, unitary representations.}
% 

\section{Introduction}  % \label{sec:intro}

% While symbolic approaches have shown promise in areas like natural language processing (NLP) and knowledge representation, they often rely on complex representations and domain-specific assumptions. For instance, in NLP,  the richness of vocabulary and context-dependent meaning can hinder the isolation of core symbolic reasoning abilities. Similarly, logic programming frequently depends on predefined relationships between symbols or hand-crafted rules, which might limit the model's capacity to uncover abstract patterns directly from the data.}

Symbolic reasoning is fundamental to diverse areas such as knowledge representation, theorem proving, and natural language processing. While large-scale Transformer models show promise in tackling complex tasks within these domains, the complexity of the models and problem settings often obscures the underlying mechanisms. In this work, we adopt a minimalist approach, focusing on a simplified setting that enables detailed analysis of how models acquire and process symbolic relationships.
% we focus on a minimal setting that {\green admits/enables} detailed analysis of how models {\red learn to process} symbolic relationships.

Symbolic operation completion (SOC) provides such a minimal setting, entailing the completion of "multiplication" tables of abstract symbols governed by binary operations (Figure~\ref{fig:cayley_tables}). This focus on basic symbolic operations isolates a core aspect of symbolic reasoning: the ability to infer relationships between symbols based on their observed interactions. SOC shares {\cyan conceptual} similarities with matrix completion, which has been instrumental in exploring theoretical questions about generalization bounds, learnability, and implicit biases in deep learning. Likewise, SOC offers the potential to unveil the fundamental principles that govern symbolic reasoning.

We demonstrate that SOC can be effectively solved %and analyzed using
by a minimal model: a bilinear map.  %a minimalist bilinear model. 
Inspired by group representation theory, 
{\blue 
we employ a novel architecture that leverages matrix embedding of symbols,  %coupled with 
and a regularizer that 
promotes learning general group structures 
---  a principle akin to the low-rank bias {\cyan used} in matrix completion.
The simplicity of this model and problem setting facilitates %allows for 
a thorough analysis, elucidating the core mechanisms underlying symbolic reasoning in SOC.
% to elucidate the core mechanism {\cyan as well as the inner representation} underlying symbolic reasoning in SOC.}
}

\section{Background}

\subsection{Low-rank Matrix Completion} %Factorization}
\label{sec:deep_matrix_factorization}

Matrix completion, the task of recovering missing entries within a matrix, is a fundamental problem with broad applications in recommender systems, data imputation, compressed sensing, and signal processing.  
Classical approaches often rely on a low-rank structural assumption, achieved through explicit rank constraints 
\citep{burerNonlinearProgrammingAlgorithm2003} 
or by minimizing the nuclear norm as a convex surrogate for rank 
% convex optimization techniques such as nuclear norm minimization 
\citep{fazelRankMinimizationHeuristic2001,candesExactMatrixCompletion2009,rechtGuaranteedMinimumRankSolutions2010,candesPowerConvexRelaxation2010}. 
Recent works have demonstrated that deep matrix factorization, %linear neural networks, 
when regularized with L2 regularization or initialized with small weights, exhibit an implicit bias towards low-rank solutions 
\citep{srebroMaximumMarginMatrixFactorization2004,gunasekarImplicitRegularizationMatrix2017}.  % saxeExactSolutionsNonlinear2014, 
This implicit approach has demonstrated improved performance in matrix completion, particularly in the limited data regime 
\citep{aroraImplicitRegularizationDeep2019a}.

% The concept of incoherence, which quantifies the spread of information within a matrix, plays a crucial role in these analyses.

\subsection{Symbolic Operation Completion (SOC)}

\citet{powerGrokkingGeneralizationOverfitting2022}  introduced SOC as a simplified setting to investigate how deep learning models, 
particularly Transformers, acquire symbolic relationships from limited data.
However, 
while Transformers could eventually solve the tasks {\cyan with extensive hyperparameter tuning}, they often struggled to generalize efficiently, requiring training times far exceeding those of simple memorization --- a phenomenon termed {\it grokking}.
Subsequent studies have further explored  this phenomenon \citep{liuUnderstandingGrokkingEffective2022a,nandaProgressMeasuresGrokking2022, chughtaiNeuralNetworksLearn2023}.

These findings suggest that Transformers may lack the appropriate inductive biases for effectively discovering structures within abstract symbolic relationships, raising broader theoretical questions about the nature of these structures and the types of biases that facilitate their discovery.
In this work, we leverage SOC as a testbed to address these questions. 
% as well as highlighting the need to identify the ideal inductive biases for SOC. %(symbolic reasoning). % and to determine whether such biases can be effectively incorporated into existing model architectures.

% A key challenge in SOC is the absence of a well-defined  {\it complexity metric} for symbolic operations, {\green akin/analogous} to rank or nuclear norm used in matrix completion. While Transformer models seem to exhibit some form of simplicity bias, their precise nature and relevance to SOC remain unclear. In this work, we propose a novel approach
% % model architecture and a regularizer
% % that imposes an implicit bias towards 
% that implicitly define a complexity metric that favors 
% learning general group structures,
% which is highly effective for solving SOC tasks.
% This approach, as we demonstrate, proves to be
% which efficiently solves SOC tasks.

\begin{figure}[t]
  \vskip 0.05in
  \begin{center}
    \hspace*{\fill}
    \includegraphics[width=0.23\textwidth]{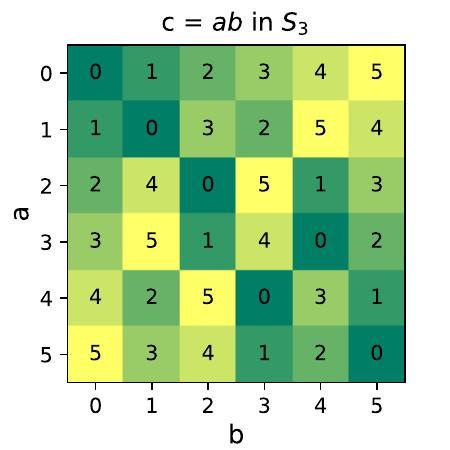}      
    \hspace*{\fill}      
    \includegraphics[width=0.23\textwidth]{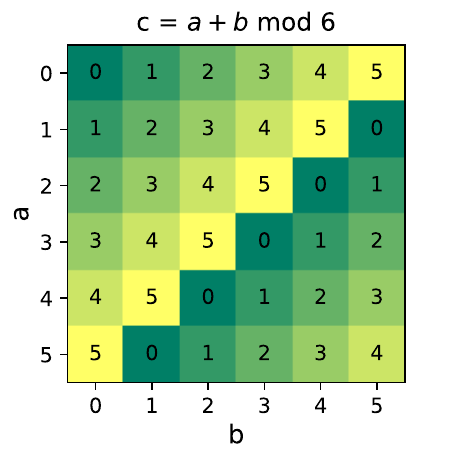}
    \hspace*{\fill}
    \includegraphics[width=0.23\textwidth]{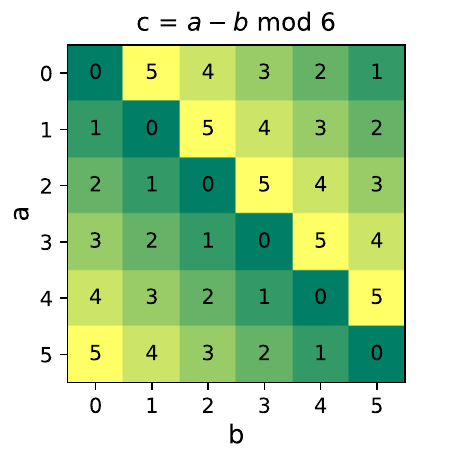}
    \hspace*{\fill}
    \includegraphics[width=0.23\textwidth]{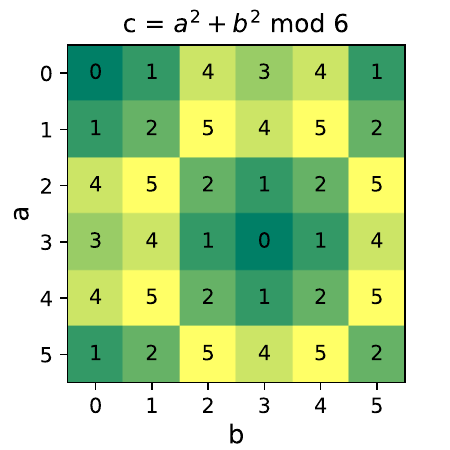}
    \hspace*{\fill}
  \caption{
    % {Small Cayley table} examples:
    % Examples of 
    Small symbolic operation tables {\cyan (Cayley tables)}:
    Symmetric (permutation) group $S_3$, modular addition, subtraction, and squared addition. 
  Elements of $S_3$ are illustrated in Figure~\ref{fig:S3_elements_illustrated}.}
  \label{fig:cayley_tables}
  \end{center}
  \vskip -0.2in
\end{figure}

\section{Group Representation Theory}

We briefly summarize group representation theory, providing key relevant concepts for our work.

\paragraph{Groups}
\label{sec:groups}
A group $(G,\circ)$ is a set $G$ with a binary operation $\circ$
that satisfies the following axioms:  
    Closure:      
    $\forall a, b \in G$, $a \circ b \in G$. 
    Associativity:      
    $(a\circ b) \circ c = a\circ (b \circ c)$. 
    Identity $e$:     
    % There exists an identity element $e$ such that   
    $g\circ e = e \circ g = g$, ${\cyan \exists e,} \forall g \in G$. 
    Inverse:   
    For every $g$, there exists a {\blue unique} inverse $g\inverse$ such that 
    $g \circ g\inverse = g\inverse  \circ g = e$.

% {\cyan A symmetry group is a group whose elements are functions, called transformations, and the operator is function composition.}

\paragraph{Representations}

A representation of a group $(G,\circ)$ on a vector space $V$ is a {\it group homomorphism} $\varrho \colon G\to \mathrm {GL} \left(V\right)$
that preserves the group structure:   
{\it i.e.}
\begin{align}
   \label{eq:group_homomorphism}
   ~~~~~~~ ~~~~~~~  ~~~~~~~
   \varrho (g_{1} \circ g_{2})=\varrho (g_{1})\varrho (g_{2})   ,  ~~~~~~~~~~ \forall g_{1},g_{2}\in G.   
\end{align}
For a vector space of finite dimension $n$, we can choose a basis and identify $GL(V)$ with $GL(n, K)$, %, \RC)$, 
{\it i.e.} the group of $n \times n$ invertible matrices over the field $K$.

\paragraph{Equivalent Representations}
Given two vector spaces $V$ and $W$, two representations $\varrho \colon G\to \mathrm {GL} \left(V\right)$ and $\pi \colon G\to \mathrm {GL} \left(W\right)$ are equivalent % isomorphic 
if there exists a vector space isomorphism $M : V \to W$, 
such that for all $g \in G$,
% \begin{align}
%     \label{eq:similarity_transform}
  $  M \varrho (g) M\inverse =\pi (g),$
% \end{align}
{\it i.e.} a similarity transformation.
% {which is called a similarity transformation.}

\paragraph{Unitary Representations}
A representation $\varrho$ of a group $(G,\circ)$ is considered {\it unitary} if every matrix $\varrho (g)$ is unitary for all $g \in G$.  Crucially, the {\it Unitarity Theorem} guarantees that any finite-dimensional representation of a compact/finite group can be expressed as an equivalent unitary representation.

\paragraph{Irreducible Representations}
A representation is considered {\it reducible} if it can be decomposed into a direct sum of smaller representations via a similarity transform,  % if a similarity transform can decompose it into a direct sum of smaller representations. 
% This decomposition 
which leads to a block-diagonal matrix form where each block corresponds to a simpler representation. 
% In contrast, 
{\it Irreducible} representations ({\it irreps})
% (often abbreviated as {\it irreps}) cannot be broken down further. They 
serve as the fundamental building blocks for constructing all possible group representations.
% {\it Irreducible} representations (often abbreviated as {\it irreps}) are the simpler representations that cannot be broken down further, forming the fundamental building blocks for constructing all possible group representations.

\paragraph{Regular Representation}
Every group $(G,\circ)$ possesses an inherent action on itself 
% through translations. This internal operation 
that can be viewed as a permutation, where each group element rearranges the other elements. 
The {\it regular} representation uses the permutation's basis vectors to construct a linear representation.
% Remarkably, the regular representation 
It is decomposible  into a direct sum of the \emph{complete} set of irreps, 
where each irrep appears with a multiplicity equal to its dimension. 
% An important property of the regular representation is that 
Moreover, its trace,
% that the trace of representation, 
also known as {\it character}, 
is a simple function:
\begin{align}
   \label{eq:trace_of_regular_representation}
    \trace{\varrho}(g) = 
      %  \begin{cases}
        n ~~ \text{if } g=e  
        ~~ \text{else} ~~ 0 . 
    % \end{cases}
\end{align} %if $g \neq e$.

\paragraph{Real vs Complex Representations}

Complex representations ($K=\mathbb{C}$) provide a rich mathematical framework for analyzing group structures in representation theory. We utilize this framework to establish the theoretical foundations of our approach in Sections~\ref{sec:model_and_method} and \ref{sec:analysis}.
However, for finite groups, real representations ($K=\mathbb{R}$) often suffice in practice,\footnotemark offering advantages in implementation and visualization.  Our empirical results in Sections \ref{sec:toy_results} and \ref{sec:grok_results} thus utilize real representations. 
% \cyan while our theoretical results in the complex domain remain directly applicable. 

\footnotetext{For finite groups, every complex representation can be realized over the real numbers with a doubling of the dimension.}

% We discuss complex diagonal representations for commutative groups in the appendix.

\section{Modeling Framework} %{Model} 
\label{sec:model_and_method}

\paragraph{Notations and Definitions}
We use the following capital symbols for order-3 tensor factors: %{\it e.g.} 
$A,B,C$. 
% {\green $T_{abc}$} denotes the entry of the tensor $T$ at index $(a,b,c)$. %$T_i$ denotes the matrix slice of $T$ at the first index $i$, and $T_{\cdot \cdot i}$ is the slice at the third index.
$A_a$ denotes the matrix slice of $A$ at the first index $a$
and $A_a\trans$ denotes its {\cyan conjugate} transpose. % of $A_a$.
$A_a B_b$ denotes the matrix product of $A_a$ and $B_b$.
% We employ the Einstein summation convention, 
Einstein convention is used, where repeated indices implies a contraction, 
{\it i.e.} summation over the index:
{\it e.g.} $A_a A_a\trans \equiv \sum_{a} A_{a} A_{a}\trans $,
unless noted otherwise.
%  e.g. in eq~\eqref{eq:matrix_slice_unitarity}.
% 
% We introduce the following  notation for gradient 
% $\nabla_{A_a} \mathcal L  \equiv \partial \mathcal L / \partial{A_a} $,
% % $\nabla_{B_b}  \mathcal L \equiv \partial \mathcal L / \partial{B_b} $,
% % $\nabla_{C_c}  \mathcal L \equiv \partial \mathcal L / \partial{C_c} $,
% $\nabla_{T_{abc}}  \mathcal L  \equiv \partial \mathcal L / \partial{T_{abc}}$.

\subsection{Symbolic Operations as Bilinear Maps}
\label{sec:modeling_framework}

% We model 
Consider a binary operation $\circ: S \times S \rightarrow S$  over a finite set $S$ containing $n$ distinct symbols: {\it i.e.} 
$ a \circ b = c$, %$ \circ(a,b) = c$, 
where $a,b,c \in S$.
To facilitate modeling, we linearize the problem by considering %/taking} 
a homomorphism $\phi:(S, \circ) \rightarrow (V, \mathcal D)$, 
where $V$ is a vector space and  $\mathcal D: V \times V \rightarrow V$ is a bilinear map over $V$,
such that 
$ \mathcal D (\phi(a), \phi(b)) = \phi(a \circ b)$.
% $ \phi(a \circ b) = \mathcal D (\phi(a), \phi(b))$.
% 
% This establishes a correspondence: $ \mathcal D (v^a, v^b) = v^c$, where $v^a,v^b,v^c \in V$ represent the symbols $a,b,c$, respectively.
% 
Specifically, we use the vector space $V = \RC^n$ with a standard basis, 
{\it i.e.} encoding each symbol as a one-hot vector.
In this framework, the bilinear map $ \mathcal D$ is represented by 
an order-3 tensor $D \in \RC^{n \times n\times n}$, % $n \times n\times n$ tensor $D$,
% the data tensor, 
whose entries are 
\begin{align*}
  D_{abc} = 
   \begin{cases}
    % 1 &\text{if  $ \circ(a,b) = c$} \\ 
    1 &\text{if  $ a \circ b = c$} \\ 
    0 &\text{otherwise},
      \end{cases}
\end{align*}
% \newline
where %Note that
elements of $S$ are used as indices of $D$ for easy readability.

This framework shows that  any binary operation over a finite set 
% $\mathcal D$ over a finite set $S$ 
can be fully expressed as a tensor.
%{\cyan sparse} tensor $D$. 
Crucially, % More importantly, 
it transforms the problem of SOC % learning symbolic operations from incomplete data 
into a tensor completion problem, where we recover the missing entries of $D$ from the observed entries in the training set $\Omega_\text{train}$.

\begin{figure}[t]
  \begin{center}
  %   \vspace*{-0.5 cm}
  \includegraphics[width=0.99\textwidth]{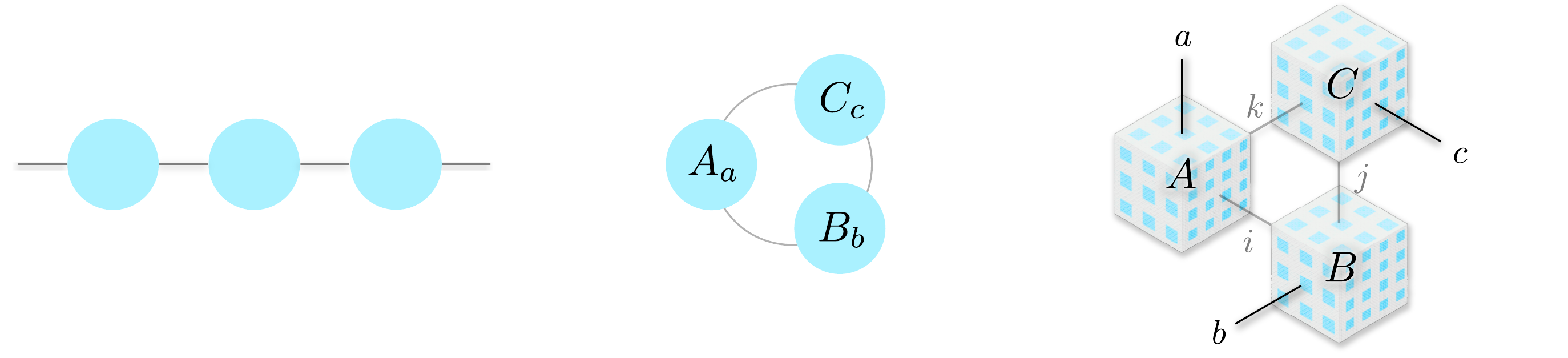}
\end{center}
  \vspace{-0.2cm}
    \caption{Visual illustration of matrix and tensor products.
    Nodes are factors and edges are indices.  
    % Internal edges (dashed) are contracted. % and the external edges (solid lines) %$a,b,c$   are exposed. % as the input/output {channels}.
    (Left) Matrix product.  (Middle) Matrix product with trace operation.  (Right)  HyperCube product.   
    % (Left) HyperCube product.      (Middle) Matrix product with trace operation.       (Right)  Matrix product.
              } 
  % \vspace{-1.0cm}
  \label{fig:HyperCube}
\end{figure}

%%%%%%%%%%%%%%%%%%

\subsection{HyperCube Parameterization} %Factorization}

We consider a minimal end-to-end model, 
a bilinear map $\mathcal{T}: V \times V \rightarrow V$, to approximate $\mathcal D$.  % $T \in \RC^{n \times n\times n}$, 
% However, directly optimizing $T$ leads to poor generalization because it treats individual entries of $T$ as independent, failing to capture the essential relationships between the entries within the tensor. To address this, 
We introduce a novel architecture %parameterization 
called {\it HyperCube}, which parameterizes the model tensor as a product of three order-3 factors $A,B,C \in \RC^{n\times n \times n}$
({\it i.e. cubes}. See Figure~\ref{fig:HyperCube}.): %{\it i.e.}
\begin{align}
  \label{eq:HyperCube_architecture}
   T_{abc}
   = \frac 1 {n}  \trace[A_a B_b C_c] 
   = \frac 1 {n} \sum_{ijk} A_{aki} B_{bij} C_{cjk}.
   % = \frac 1 {n}  \trace[A_a B_b C_c] 
   % \footnotemark
\end{align}
% \footnotetext{$T_{ab}^c$}
% 
HyperCube can be viewed as a type of tensor decomposition,%
\footnote{It is closely related to the architecture used in tensor ring decomposition \citep{zhaoTensorRingDecomposition2016}.}
but it crucially differs from existing %tensor 
decomposition methods, which often employ lower-order factors (e.g., vectors or matrices) to reduce model complexity.
% {\cyan ({\it e.g.} vector factors in CP decomposition and matrix factors in \citep{tuckerMathematicalNotesThreemode1966a})},
% It also differs from  \citep{mesnerAssociationSchemesTriples1990}, another ternary decomposition.
In contrast, HyperCube preserves the full expressive capacity of $T$ {\blue without restricting the model's hypothesis space}. 

{\blue
More intuitively, HyperCube can be understood as modeling symbols and their interactions using matrix embeddings and multiplications: % {\cyan without any nonlinearities}:
Factors $A$ and $B$ function as embedding dictionaries that map each symbol $a$ and $b$ to their respective matrix embeddings ($A_a$, $B_b$). The model then calculates the interaction between $a$ and $b$ via matrix multiplication ($A_a B_b$). Finally, factor $C$ maps this result back to the space of symbols --- {\it i.e.} the {\it unembedding} dictionary.
Importantly, this (un)embedding process is directly related to the generalized (inverse) Fourier transform on groups. See Appendix~\ref{Appendix:fourier_transform_HyperCube}.}

%%%%%%%%%%%%%%%%

\subsection{Regularization}

The model is trained by minimizing the following regularized objective:
\begin{align}
  \label{eq:regularized_loss}
  \mathcal L = \mathcal L_o(T; D)  +\epsilon \mathcal H(A,B,C) , 
\end{align}
where 
$\mathcal H$ is the HyperCube regularizer %defined on the factors as %, 
defined as 
\begin{align}
  \label{eq:HyperCube_regularizer}
  \mathcal H  \equiv   \frac 1 {n}  
  \trace \left [       A_{a}\trans A_{a} B_{b} B_{b}\trans
                    +  B_{b}\trans B_{b} C_{c} C_{c}\trans
                    +  C_{c}\trans C_{c} A_{a} A_{a}\trans    \right ] .
\end{align}
and $\mathcal L_o$ is a differentiable loss %function % defined 
on the end-to-end model, % $T$, 
% specifically % Here, we use 
{\it e.g.}  total squared error over $\Omega_\text{train}$ %the training set 
\begin{align}
  \label{eq:loss_square}
  \mathcal L_o (T; D) = 
  \sum_{(a,b,c) \in \Omega_\text{train}} 
  (T_{abc} - D_{abc})^2,
\end{align}
% 
% 
% Note that the HyperCube regularizer can also be understood as the L2 regularization on factor pairs:
% $A_{a} B_{b}$, $B_{b} C_{c}$, and $C_{c} A_{a}$.
% {\it i.e.} $\trace [  A_{a}\trans A_{a} B_{b} B_{b}\trans] = $

\subsection{Internal Symmetry of Model}
\label{sec:internal_symmetry_basis_change}

The {redundant parameterization} of eq~\eqref{eq:HyperCube_architecture}
implies the existence of 
% that there are multiple degrees of 
internal symmetry that leaves the model unchanged.  
For example, one can introduce arbitrary invertible matrices $M_I, M_J, M_K$ and their inverses between the factors as 
$\tilde{A}_a = M_K\inverse A_a M_I$, $\tilde{B}_b = M_I\inverse B_b M_J$, and $\tilde{C}_c = M_J\inverse C_c M_K$.
These yield an equivalent parameterization of $T$, since 
% all $M$'s get %the $M$ matrices all cancel out when computing the model: %end-to-end model $T$: 
% {\it i.e.} 
$\trace [\tilde{A}_a \tilde{B}_b \tilde{C}_c] = \trace[A_a B_b C_c] $.
These symmetry transformations can be understood as changing the {internal} basis coordinates. % for representing the factors. 

Note that while  the model loss $\mathcal L_o (T)$ is invariant under such coordinate changes, the regularizer $\mathcal H(A,B,C)$ is not. However, the regularizer is invariant under {\it unitary} basis changes, in which the introduced matrices are unitary $U_I, U_J, U_K$,  such that  $ U U\trans = U\trans U = I $. 
{Therefore, the regularizer imposes a stricter form of symmetry.}
This leads to the following Proposition. 
{\red 
\begin{proposition}
  If $A,B,C$ form the optimal solution of the regularized loss eq~\eqref{eq:regularized_loss},
  then  any unitary basis changes leave the solution optimal,
  but non-unitary basis changes {generally increase the loss}.
\end{proposition}
}

\section{Analyzing HyperCube's Inductive Bias}
\label{sec:analysis}

While HyperCube eq~\eqref{eq:HyperCube_architecture} does not explicitly restrict the model's hypothesis space,  the regularizer eq~\eqref{eq:HyperCube_regularizer} 
induces a strong implicit bias
% shaping the model's learning behavior and 
% that steers the model. 
that guides the model towards specific solutions. 
% that guides the model %'s convergence to specific types of solutions. 
% it exhibits a strong {\it implicit bias} when combined with the regularizer eq~\eqref{eq:HyperCube_regularizer}.
In this section, we introduce key concepts for %measuring and 
analyzing this inductive bias.
% Proofs are in Appendix~\ref{Appendix:proofs}.
See Appendix~\ref{Appendix:proofs} for proofs.

% that enable us to measure and analyze this inductive bias.
% 
% \begin{definition}
%   \label{def:imbalances}
%   We define \emph{imbalances} across edge $i,j,k$  as 
% \end{definition}

\begin{lemma}[Balanced Condition]
  \label{lemma:balanced_condition}
  At stationary points of %the regularized loss 
  eq~\eqref{eq:regularized_loss}, imbalance terms vanish to zero:
\begin{align}
  \label{eq:balanced_condition}
  \xi_I= \xi_J= \xi_K= 0,
\end{align}
where      
$\xi_I =  A_a\trans (C_{c}\trans C_{c}) A_a   - B_{b} (C_{c} C_{c}\trans) B_b\trans $, 
$\xi_J  =  B_b\trans (A_{a}\trans A_{a}) B_b  - C_{c} (A_{a} A_{a}\trans) C_c\trans$, 
and
$\xi_K  = C_c\trans (B_{b}\trans B_{b}) C_c  -  A_{a} (B_{b} B_{b}\trans) A_a\trans$
are the imbalances across edge $i,j$, and $k$, respectively.
\end{lemma}
% 
% \begin{proof}
%   % This is proved by showing that 
%   %  the fact that the gradient is zero at stationary points.
%   See Appendix~\ref{Appendix:balanced} for detailed proof.
% \end{proof}
% 
% {\cyan This result generalizes the balanced condition of deep linear neural networks under L2 regularization
% \cite{aroraOptimizationDeepNetworks2018,saxeMathematicalTheorySemantic2019,huhCurvaturecorrectedLearningDynamics2020}.}

The following statements {\red  demonstrate that the regularizer %eq~\eqref{eq:HyperCube_regularizer} 
promotes a unitarity condition}.
% {\red a form of unitarity}
% C-unitarity 
% in the factors:
% 
\begin{definition}[Contracted Unitarity] %Internal} Unitarity
  A factor $A$ is {\it C-unitary}  if it satisfies the following:
  % \begin{align}
  %   \label{eq:contracted_unitarity0}
  $  A_{a} A_{a}\trans, \, A_{a}\trans A_{a}      \propto I $
  % \end{align}
  (\emph{with} contracting the repeated index $a$).
\end{definition}

\begin{proposition}
  \label{proposition:contracted_unitarity}
  C-unitary factors satisfy the balanced condition eq~\eqref{eq:balanced_condition}, given that they share a common scalar multiple of the identity matrix: {\it i.e.}
\begin{align}
  \label{eq:contracted_unitarity2}
  A_{a} A_{a}\trans = A_{a}\trans A_{a} 
  & =  B_{b} B_{b}\trans   = B_{b}\trans B_{b}   %\nonumber \\  
   = C_{c}  C_{c}\trans  = C_{c}\trans C_{c} 
  \equiv n \alpha^2 I ,
\end{align}
\end{proposition}

\begin{lemma} %[{Stationarity} of Unitary Factors]
  \label{lemma:Frobenius_fixed_regularizer}
  Under the fixed Frobenius norm, %eq~\eqref{eq:L2_regularizer}, 
  all C-unitary factors are stationary points of the regularizer $\mathcal H$.
\end{lemma}

% \begin{proof}
%   Proposition~\ref*{proposition:contracted_unitarity}
%   can be confirmed in a straightforward manner.
%   Proof for Lemma~\ref{lemma:Frobenius_fixed_regularizer} is in Appendix~\ref{appendix:Frobenius_fixed_regularizer}.
% \end{proof}

{\cyan {Lemma}~\ref{lemma:Frobenius_fixed_regularizer}  indicates that $\mathcal H$ effectively promotes C-unitarity 
as well as %while also 
minimizing the Frobenius norm.}
% % 
% 
% {\red Remarkably, we observe that the converged solutions  
% % the model converges solutions that 
% can exhibit  a stronger form of unitarity.}
Remarkably, we also observe  a stronger form of unitarity in the converged solutions.
% the model converges solutions that 

\begin{definition}[{Slice} Unitarity]
  A factor $A$ is \emph{S-unitary} if every matrix slice of $A$ is a scalar multiple of an unitary matrix:
  {\it i.e.} 
  $  A_{a} A_{a}\trans = A_{a}\trans A_{a} \equiv \alpha_{A_a}^2 I $
(\emph{without} contracting the repeated index $a$).
\end{definition}
% 
% Note that S-unitarity is a stronger and more detailed condition than C-unitarity.  
% Remarkably, we make the following observation:
% 
% 
\begin{observation}
  \label{observation:matrix_slice_unitarity_always}
  When optimizing the regularized loss eq~\eqref{eq:regularized_loss}, C-unitary solutions  are {consistently} achieved via S-unitarity, 
  % in which case it %In this case, eq~\eqref{eq:contracted_unitarity2} 
  in which eq~\eqref{eq:contracted_unitarity2} reduces to
  % reduces to
{\small $\sum_a \alpha_{A_a}^2 = \sum_b \alpha_{B_b}^2 = \sum_c \alpha_{C_c}^2 = n\alpha^2$}.
\end{observation}

{\red Although the exact mechanism driving S-unitarity remains an open question, this observation highlights the strong inductive bias towards unitarity imposed by the HyperCube regularizer.}

\section{Representation Learning in HyperCube} %Small Operation Datasets}
% \section{Small Dataset Experiments}
\label{sec:toy_results}

% These small-scale experiments offer several advantages: they enable close examination of the model's behavior, full visualization of its learned representations, and ultimately, a deeper understanding of the model's key operating mechanisms.

\subsection{Learning Dynamics on partially observed $S_3$}
% \label{sec:toy_learning_dynamics}

\begin{wrapfigure}{r}{0.58\textwidth}  
  \begin{center}
    \vspace*{-1 cm}
    % \vspace*{-1.2 cm}
    % \vspace*{-0.8 cm}
    \includegraphics[width=0.58\textwidth]{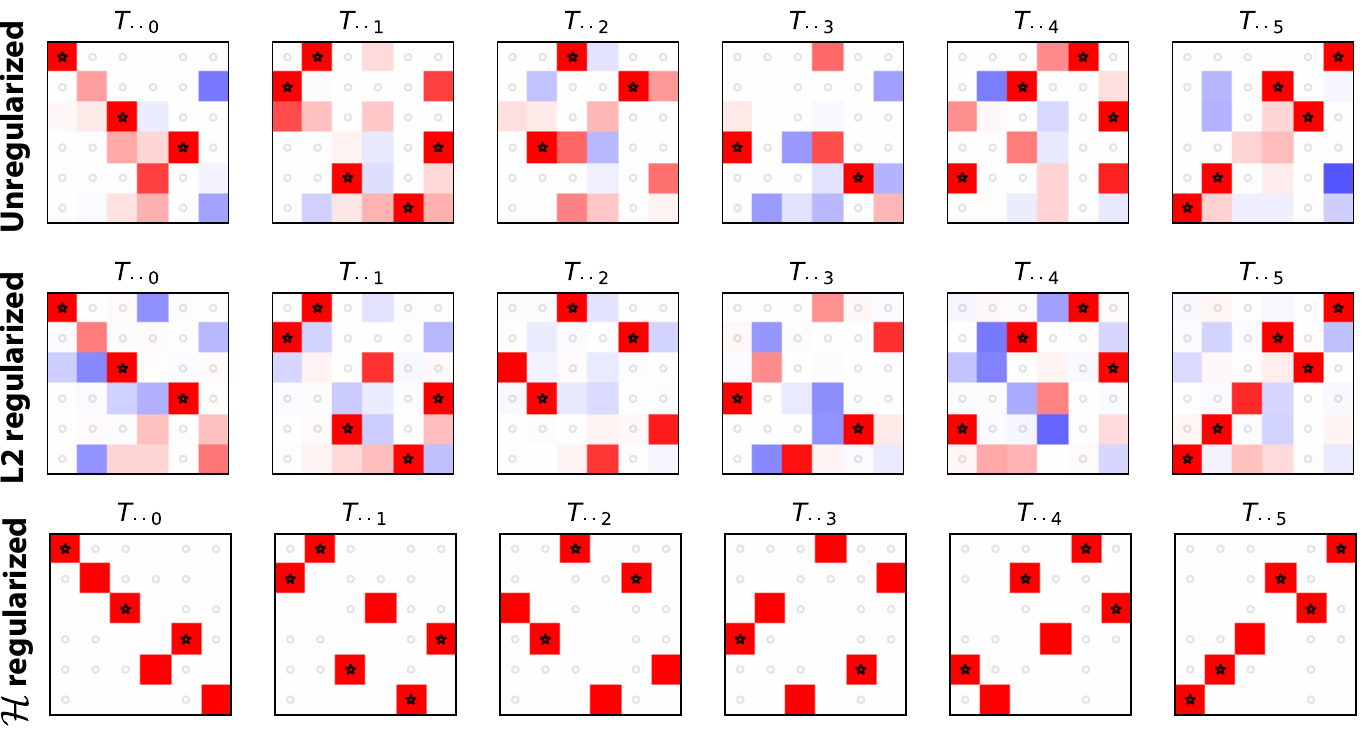}%
    \vspace*{-0.1 cm}
    % \vspace*{-0.25 cm}
  \caption{   Model slices $T_{\cdot\cdot c}$ after trained on the $S_3$ dataset.
    Training data  are marked by stars (1s) and circles (0s). %$\Omega_\text{train}$ 
    % (Color scheme: red=1, white=0, blue=-1.)
    % A similar visualization of the modular addition experiment is shown in Figure~\ref{fig:add6_T}.
  }
  % \vspace*{-0.4 cm}
  \vspace*{0.8 cm}
  \label{fig:sym3_T}
  \end{center}
\end{wrapfigure}

\begin{figure}[t]
  % \vskip 0.05in
  \begin{center}
    \vspace{-1em}
    \includegraphics[width=0.99\textwidth]{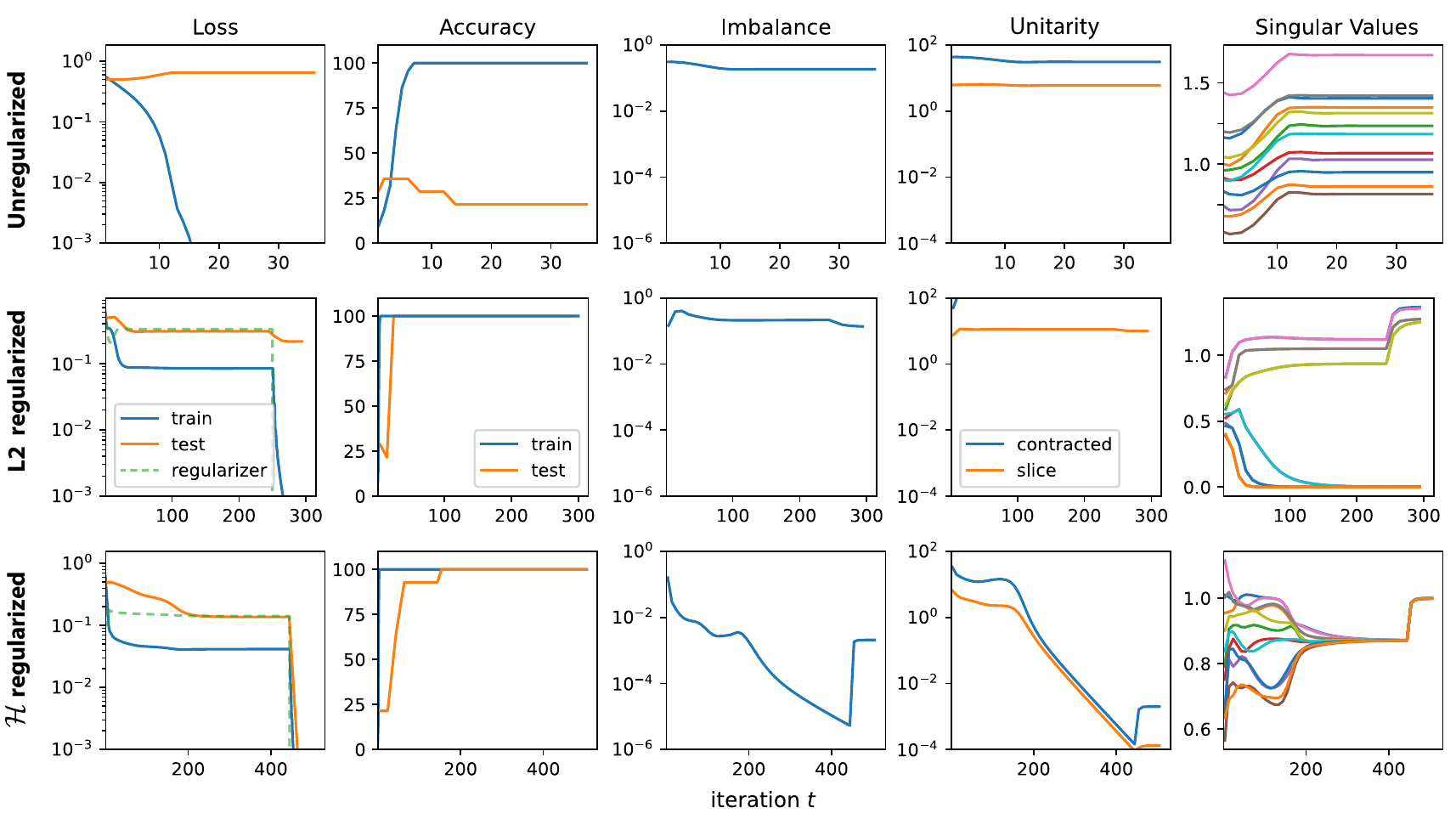}

    \vspace{-1em}
  \caption{
    Optimization trajectories on the %symmetric group 
    $S_3$ dataset with 60\% training data fraction. %of Cayley table used as the training set (See Figure~\ref{fig:sym3_T}).
    (Top) Unregularized,       (Middle) L2-regularized, and       (Bottom) $\mathcal H$-regularized training. 
    Column 3 shows the average imbalance       $(\Vert \xi_I\Vert^2_F +\Vert\xi_J\Vert^2_F + \Vert\xi_K\Vert^2_F)^{1/2}$,
    and column 4 shows  deviation from C-unitarity 
    $\Vert \sum_a A_{a} A_{a}\trans/n - \alpha^2 I \Vert^2_F$
    and S-unitarity 
    % \eqref{eq:matrix_slice_unitarity},
    $\Vert A_{a} A_{a}\trans - \alpha_{A_a}^2 I \Vert^2_F$,      averaged over all factors and slices. 
    Column 5 shows  normalized singular values of unfolded %matrices obtained by unfolding 
    factors $A,B,C$. 
    % In the $\mathcal{H}$-regularized case, the  singular values converge to a common value,     corresponding to $\alpha$ in eq~\eqref{eq:contracted_unitarity2}.
    % Only a subset is shown for clarity. 
    }
  \label{fig:sym3_training_traj}
  \end{center}
  \vskip -0.1in
\end{figure}

We begin our analysis by examining how our model learns
% We first analyze how our model learns %its behavior on 
the symmetric group $S_3$ (using 60\% of Cayley table as training data).  Figure~\ref{fig:sym3_training_traj} visualizes the optimization trajectories under different regularization strategies, while Figure~\ref{fig:sym3_T} depicts the resulting end-to-end model. 
The evolution of the model and its parameters is directly visualized in Figure~\ref{fig:sym3_weight_hist}.
See Appendix~\ref{appendix:Training_Procedure} for training details.

In the absence of regularization, the model quickly memorizes the training dataset, achieving perfect training accuracy, but fails to generalize to the test dataset. Also, the singular values of the unfolded factors remain largely unchanged during training, indicating minimal internal structural changes.

% \paragraph{$\mathcal H$ regularized model}
Under HyperCube ($\mathcal H$) regularization, 
the model continues to improve on the test set even after perfect training accuracy is achieved. 
% 
% {\red the model memorizes the train dataset just as rapidly, but %{\cyan then} it} continues to improve the test set performance. % even long after the initial memorization phase. 
% 
A critical turning point is observed around $t= 200$, %(vertical dashed line),
marked by a sudden collapse of the singular values towards a common value, signifying convergence to a unitary solution. 
% From this point onward, 
Concurrently, the C/S-unitarity and imbalance measures rapidly decrease to zero. This internal restructuring coincides with a substantial improvement in test performance, ultimately achieving 100\% test accuracy, 
thus highlighting its crucial role in enabling generalization.
Notably, when the regularization coefficient drops to 0 % $\epsilon$-scheduler reduces $\epsilon \to 0$ 
around $t=450$, both the train and test losses plummet to 0, confirming perfect completion %recovery 
of $D$.

% {\red This sudden convergence to unitarity is also evident in Figure~\ref{fig:sym3_weight_hist}, which directly illustrates the product and factor tensors during the learning process.  In stark contrast, the factors of the unregularized model remain largely unchanged from their random initial values.}

% \paragraph{L2 regularized model}    
In contrast, L2 regularization drives the model towards a low-rank solution, as evidenced by a portion of its singular values decaying to zero.
Although this training scheme shows some degree of generalization, it fails to reduce the test loss to zero, indicating imperfect recovery. {\blue 
On the modular addition task (Figure~\ref{fig:add6_training_traj},\ref{fig:add6_T}), L2-regularized training %{\red indeed} 
only achieves $\sim$50\% test accuracy.}
Figure~\ref{fig:sym3_T} visually confirms these findings, demonstrating 
% that neither the unregularized nor the L2-regularized models fully recover the data tensor $D$.
that only $\mathcal H$-regularized training is capable of accurately recovering $D$.

\subsection{Model Learns Unitary Group Representations}
% \subsection{Learned Factors are Unitary Group Representations}
\label{sec:factors_are_group_representations}

% This section analyzes the HyperCube factors trained on $S_3$. 
% properties of factors learned by HyperCube. 
% Figure~\ref{fig:basis_change} visualizes
In Figure~\ref{fig:basis_change}, we analyze the learned factors in different basis coordinate representations, demonstrating a remarkable finding: The factors directly encode group representations.
% as shown in Figure~\ref{fig:basis_change}.
% revealing remarkable structures:
% 
% \newline

% \textbf{Initial Observation} 
(Top panel)  In the raw basis coordinate, % representation, 
the factors exhibit unitary matrix slices, but no other easily identifiable structure.

% \textbf{Key Transformation} 
(Middle panel)  A unitary change of basis, such that the factor slices for the identity element become the identity matrix (${A}_e = {B}_e = {C}_e = I$), reveals a surprising underlying structure: 
\begin{itemize}[leftmargin=*]
    \item %\textbf{Weight Equality}: 
    All factors share the same embedding, {\it i.e.} %meaning 
    ${A}_g$ equals ${B}_g$ equals %also equals 
    ${C}_g\trans$  for all elements $g$.
    % the conjugate transpose of ${C}_g$ $\forall g \in G$. % forall % 
    \item  %\textbf{Group homomorphism}: 
    The factor slices 
    satisfy {\it group homomorphism}  eq~\eqref{eq:group_homomorphism}:
    % respect the underlying group structure. 
    % Multiplying slices corresponds to the group operation:
    {\it i.e.} ${A}_{g_1} {A}_{g_2} = {A}_{g_1 \circ g_2}$.
    See Figure~\ref{fig:AA_multiplication_table}.
    % satisfying {\it group homomorphism}  eq~\eqref{eq:group_homomorphism}.
% \end{itemize}
    \item 
    Therefore, the factors form a 
    % proper 
    unitary  matrix representation $\varrho$ of the group, where
    \begin{align}
        \label{eq:factors_are_group_representations}
        {A}_g={B}_g={C}_g\trans =  \varrho(g).
    \end{align}
    \item 
    Furthermore, the trace of factor slices satisfy eq~\eqref{eq:trace_of_regular_representation},
    indicating that $\varrho$ is a regular representation. % of the group.
\end{itemize}

% Figure~\ref{fig:AA_multiplication_table} demonstrates the group homomorphism property. The multiplication table of factor slices mirrors the structure of the group's Cayley table (Figure~\ref{fig:cayley_tables}).

% \textbf{Block-diagonalization} 
(Bottom panel)
% Representing in a block-diagonalized form 
In a block-diagonalizing basis coordinate, the factors reveal
% The basis can be further changed to block-diagonalize the factors, revealing 
the {\it complete set of irreps} contained in the regular representation $\varrho$, including the trivial (1-dim), sign (1-dim), and duplicate standard representations (2-dim),
{\cyan which form the generalized Fourier basis for group convolution.}

\paragraph{Shared-Embedding}
Eq~\eqref{eq:factors_are_group_representations} reveals that, for group operations, the same embedding is used across all symbol positions. This motivates tying the embeddings across factors, resulting in a parameter-efficient model specifically tailored for learning group operations: HyperCube-SE (shared embedding).
\vspace{-0.4cm}
\paragraph{\red Key Operating Mechanism}
Above results reveal the key mechanism by which HyperCube operates on groups.  
According to  
eq~\eqref{eq:factors_are_group_representations}
and the homomorphism property of $\varrho$, % eq~\eqref{eq:group_homomorphism}, 
the model eq~\eqref{eq:HyperCube_architecture} can be expressed as 
\begin{align}
    \label{eq:product_tensor}
    T_{abc}     =  \frac 1 n \trace [\varrho{(a)} \varrho{(b)} \varrho{(c)}\trans]
                = \frac {1} {n} \trace [\varrho(a \circ b \circ c\inverse)].
\end{align}
Since $a \circ b = c$ is equivalent to $a \circ b \circ c\inverse = e$, 
applying eq~\eqref{eq:trace_of_regular_representation}
% due to the trace condition eq~\eqref{eq:trace_of_regular_representation} of $\varrho$, 
implies that  $T_{abc}=D_{abc}$. 
Notably, this mechanism applies universally for all finite groups, 
{\cyan yielding exact one-hot encoding for the output symbols.}
This insight leads us to the following conjecture:

\begin{conjecture}
  \label{conjecture:group_rep_is_minimizer}
Let $D$ represent a group operation table. 
Then, given the constraint $T=D$,  the unitary group representation eq~\eqref{eq:factors_are_group_representations} describes the unique minimizer of HyperCube Regularizer eq~\eqref{eq:HyperCube_regularizer} up to unitary basis changes,
{\blue whose %and the 
minimum regularizer loss is 
$\mathcal{H}^*(D)  = 3 \Vert D \Vert_F^2$.} %  = 3 n^2
\end{conjecture}

\subsection{Discovering Unitary Representations Beyond True Groups}   
\label{sec:group-like}

% Besides the $S_3$ example, 
We trained HyperCube on the remaining small operation tasks from Figure~\ref{fig:cayley_tables}. % besides the $S_3$ example. 
Interestingly, the model learns closely related representations across these tasks
% {\red even for {\it group-like} operations}  
(See Figure~\ref{fig:+-quad}).

% \newline
\textbf{Modular addition} ($a + b$) forms the cyclic group $C_6$. As expected, HyperCube learns the regular representation $\varrho(g)$ of  $C_6$ in its factors, as described by eq~\eqref{eq:factors_are_group_representations}.
% {\red (See Figure~\ref{fig:add6_training_traj} and {\ref{fig:add6_T}} for more results on modular addition).}

% Next, we explore operations resembling groups but lacking certain axioms.
% 
\textbf{Modular Subtraction} ($a - b$)
% : This operation 
violates associativity and therfore isn't a true group. Surprisingly, HyperCube still learns the same representation as addition but with transposed factors: 
${A}_g\trans={B}_g={C}_g = \varrho(g)$.
This reflects % This intriguing behavior reflects % is due to
the equivalence: $a - b = c \Leftrightarrow  a = b + c$. % essentially {\it wrapping around} the group structure.

\textbf{Modular Squared Addition} ($a^2 + b^2$)
% : This operation 
violates the inverse axiom. Still, HyperCube learns the same representation as addition for elements with unique inverses (e.g., $0,3$). For others, it learns {\it duplicate} representations reflecting the periodicity of squaring modulo: % $6$:
{\it e.g.}, $A_2 = A_4$ since $2^2 = 4^2$(mod 6). % and $A_1 = A_5$, {\cyan and $1^2 = 5^2$.} 

These results highlight the remarkable flexibility of HyperCube's inductive bias: Even for {\it group-like} operations (i.e., those deviating from strict group axioms), HyperCube often discovers meaningful unitary representations. 
% This suggests that its inductive bias extends beyond classic group structures, potentially enabling the model to learn a wider range of symbolic relationships.
This finding highlights the potential of unitary representations as a powerful tool for understanding symbolic operations beyond the confines of strict group theory.

\section{Results on Diverse SOC Datasets} %Larger Dataset Experiments} 
\label{sec:grok_results}

\begin{figure}
    \begin{center}
      \includegraphics[width=0.95\textwidth]{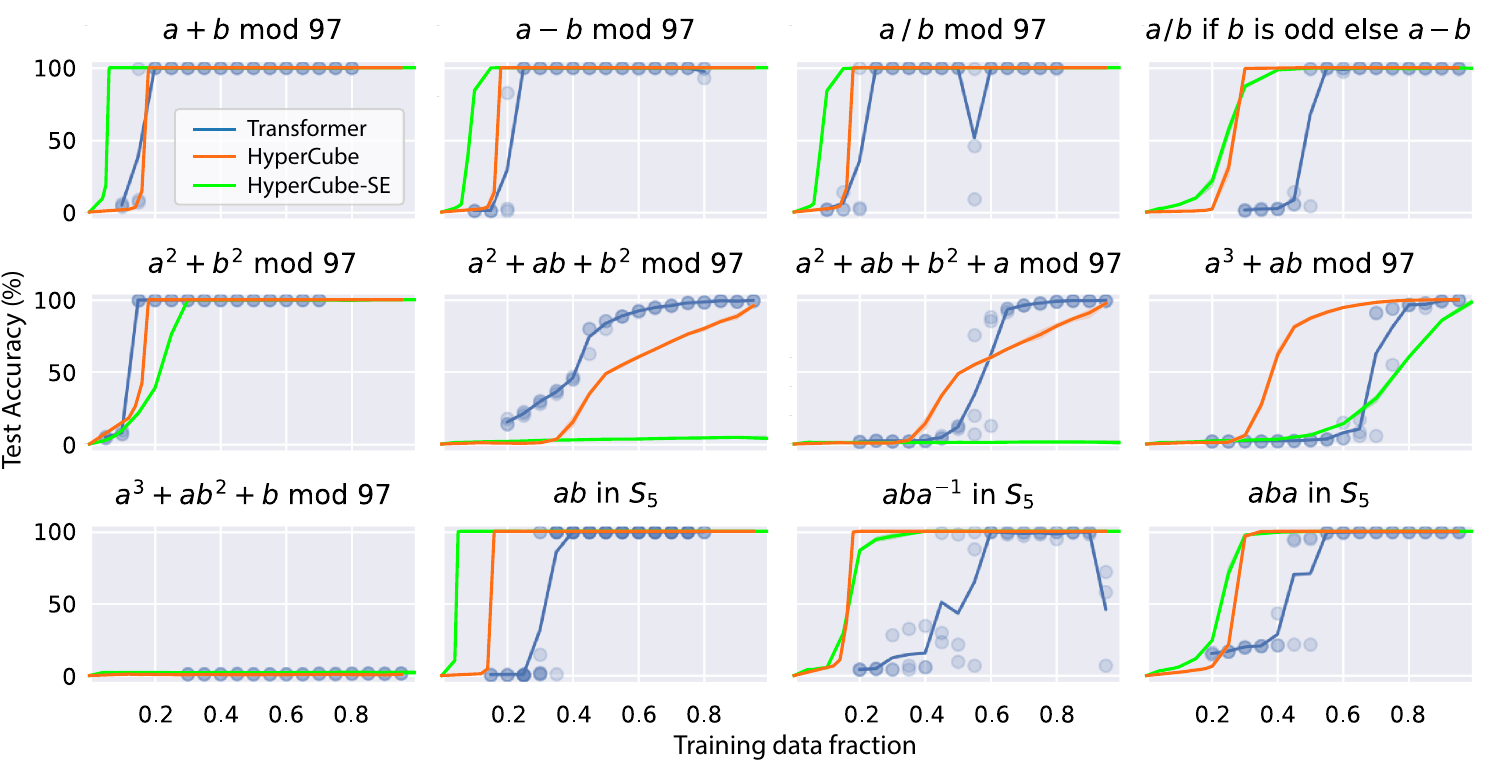}
    \end{center}
    \vspace{-0.3cm}
      \caption{
        % Comparison of 
        Generalization performance (test accuracy) shown as a function of training data fraction
        across a diverse set of symbolic operation tasks.
        % Models: Transformer,  HyperCube, and HyperCube-SE. % weights.
        Trial-to-trial variation due to randomized model initialization and data split is shown as dots for Transformer and as shaded area for HyperCubes. % and HyperCube-SE.
      %   Transformer often shows large trial-to-trial variations (dots) 
      %   % Individual trial data is shown for Transformer (dots), revealing trial-to-trial variations. 
      % whereas HyperCube and HyperCube-SE show negligible variations %across trials 
      % (shaded area).
      % due to consistent, stable learning dynamics
    } 
    \vspace{-0.1cm}
    \label{fig:grok_customL2}
  \end{figure}

We trained HyperCube and HyperCube-SE on diverse SOC datasets from \citet{powerGrokkingGeneralizationOverfitting2022},  encompassing various group and non-group operations (details in Appendix~\ref{appendix:binary_operations_list}). These problems are significantly larger than our previous examples, with dimensions ranging from $n=97$ to $120$. Figure~\ref{fig:grok_customL2} compares their performance to the baseline Transformer results from \citet{powerGrokkingGeneralizationOverfitting2022}.

HyperCube demonstrates remarkable generalization across a wide range of tasks. On {\it simpler} tasks, it achieves perfect test accuracy with only $\sim$18\% of the data, 
including group and {\it group-like} operations (Sec.~\ref{sec:group-like}) with known unitary representations, 
as well as % modular division ($a/b$) and 
group conjugation ($a \circ b \circ a\inverse$ in $S_5$), which lack such representations.
For more {\it complex} tasks, HyperCube requires more data for effective generalization, such as modular polynomials, conditional operations, and $a \circ b \circ a$ in $S_5$.
Overall, Hypercube exhibits comparable or superior generalizability  to the Transformer baseline on all but two tasks (modular $a^2+ab+b^2$ and $a^2+ab+b^2+a$).

{\blue HyperCube-SE demonstrates a similar trend but with a narrower inductive bias towards group operations. %also exhibits a similar trend but with a narrower inductive bias: % {\red towards} %specific to} group operations:
It requires even less data %($\sim 5\%$) 
for group operations but shows weaker generalization on non-group operations, especially high-order modular polynomials.}
% more data for high-order polynomials.}

\begin{figure}
\centering
\begin{minipage}{.54\textwidth}
  \centering
  \includegraphics[width=.98\linewidth]{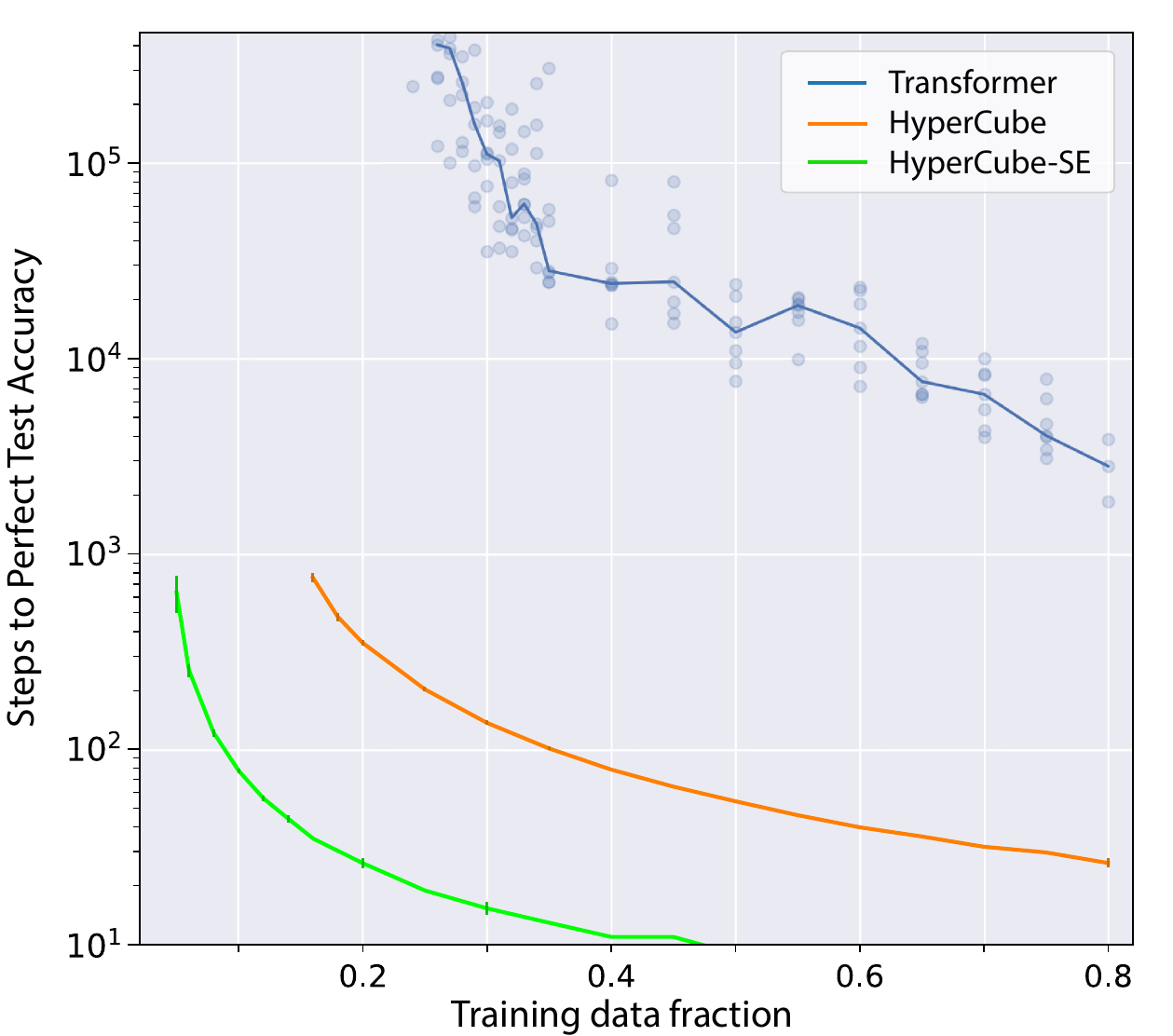}
  \caption{
          Training steps to achieve perfect test accuracy on the ($ab$ in $S_5$) task. % \cyan (with 1-sigma errorbars).  % (>99\%) %  with varying amounts of training data. 
          % Models:
          % HyperCube  (orange),
          % HyperCube with shared factor weights (green), and 
          % Transformer \citep{powerGrokkingGeneralizationOverfitting2022} (blue). 
        }
        \vspace{-0.1cm}
  \label{fig:S5_slow}
\end{minipage}%
\hfill
\begin{minipage}{.43\textwidth}
  \centering
  \includegraphics[width=.96\linewidth]{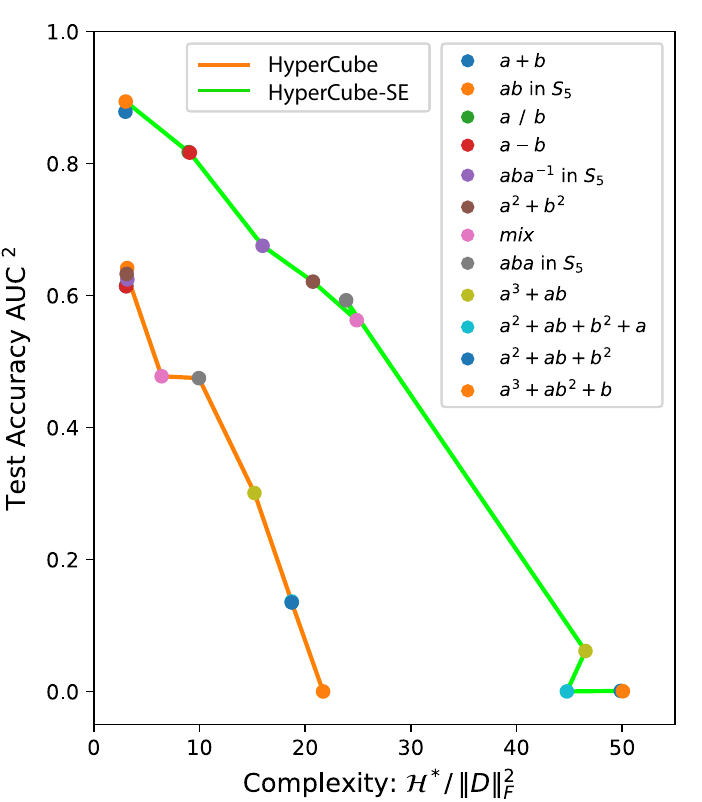}
  \caption{
    Complexity vs Generalizability. (AUC $\equiv$  Area Under the Curve)
  }
  \vspace{-0.1cm}
  \label{fig:complexity_vs_accuracyAUC}
\end{minipage}
\end{figure}

\paragraph{Blazing-Fast Learning}
Beyond its sample efficiency, HyperCube exhibits exceptional learning speed. As shown in Figure~\ref{fig:S5_slow} on the $S_5$ group operation, it converges to perfect test accuracy 100 times faster than the Transformer baseline, while requiring less data.  HyperCube-SE, with shared factor weights ($A_g=B_g=C_g\trans$), similar to shared embeddings in Transformers, achieves an additional 10$\times$ speedup and requires only $5\%$ of the data for perfect generalization. 
This dramatic 1000$\times$ improvement in learning speed demonstrates the effectiveness of our models' inductive bias.

% This represents a dramatic 1000$\times$ improvement in learning speed compared to the Transformer.

\paragraph{Complexity vs Generalizability}

{\blue  
In matrix factorization, the minimum L2 regularization loss {\green implicitly defines} a {\it complexity metric} that approximate rank:
{\it e.g.}, nuclear or Schatten norm \citep{srebroMaximumMarginMatrixFactorization2004}.
% that favors low-rank solutions: {\it e.g.} nuclear norm or Schatten norm \citep{srebroMaximumMarginMatrixFactorization2004}.
% \citep{saxeExactSolutionsNonlinear2014}
% 
% 
Similarly, 
we formally define a complexity metric for SOC
% in SOC, we can formally define a complexity metric
% The complexity metric for SOC can be formally defined by 
as the minimum HyperCube regularization $\mathcal{H}^*$ %{\green that incurs when fitting/
required to fit the full operation table. 
This metric aligns with the intuitive notion of complexity in symbolic operations (Figure~\ref{fig:complexity_vs_accuracyAUC}).
Group operations achieve the minimum complexity $\mathcal{H}^* = 3 \Vert D \Vert_F^2$,  indicating their inherent simplicity within HyperCube. {\it Group-like}  operations also achieve this minimum in HyperCube but incur increased complexity in HyperCube-SE, demonstrating the latter's narrower inductive bias towards pure group structures.   In contrast, more complex tasks, such as modular polynomials, incur substantially higher complexity costs, culminating in the unsolvable cubic operation (modular $a^3 + ab^2 + b$).
}

Figure~\ref{fig:complexity_vs_accuracyAUC} illustrates the generalization trends as a function of complexity, revealing a clear monotonic relationship: increasingly complex tasks exhibit lower generalizability (measured as the total area under the test accuracy curve).
This observation underscores the critical role of our proposed complexity metric in determining the generalization {\blue bound} for SOC,
mirroring the relationship between matrix rank, observed entries, and generalization error in matrix completion.

% {\cyan Transformer also shows a similar trend, requiring more data for complex tasks  from \citet{powerGrokkingGeneralizationOverfitting2022}.
% % 
% However, it favors commutative operations ({\it e.g.}, $a+b$, $a^2+ab+b^2$) over non-commutative ones ({\it e.g.}, $a-b$, $a^2+ab+b^2+a$, and all $S_5$ tasks) \cite{powerGrokkingGeneralizationOverfitting2022,liuUnderstandingGrokkingEffective2022a}.
% This may result from Transformers sharing symbol embeddings across input locations. %In contrast, HyperCube handles both types equally well.
% }
% 

\section{Conclusion}

In this work, we investigated symbolic operation completion (SOC) as a minimal yet fundamental task for studying symbolic reasoning. 
We demonstrated that these tasks can be effectively solved using 
a simple bilinear model with a factorized architecture, 
% analogous to approaches used in matrix completion, %thus 
% {\green establishing} %/highlighting} 
% a {\cyan novel} connection between these two seemingly disparate domains. 
% The simplicity of our model has enabled 
% and conducted a thorough analysis to 
and revealed the key principles underlying symbolic reasoning in SOC by analyzing the model. % learning process and representations.

{\blue
Our core innovation lies in representing symbols and their interactions via matrix embeddings and matrix multiplications, modeling each symbol as an operator that dynamically influences others.  
This operator-based approach%
\footnote{The operator-based perspective is also fundamental in quantum mechanics, where unitary matrix operators describe how physical systems evolve over time.}
aligns with the principles of {\it dynamic semantics} in linguistics, where words are seen as context-changing operators 
\citep{kampTheoryTruthSemantic1981,heimSemanticsDefiniteIndefinite1982a}. It  also resonates with earlier explorations of compositional representations in connectionist models \citep{smolenskyTensorProductVariable1990} and recursive neural networks \citep{socherSemanticCompositionalityRecursive2012}.
This contrasts with traditional vector embeddings ({\it e.g.}, word embeddings), which primarily capture static semantic meaning and necessitate %require 
additional mechanisms to model interactions between symbols  ({\it e.g.}, self-attention in Transformers).
% While such mechanisms can approximate dynamic semantics by contextualizing word representations, {\red they are ultimately limited by the fixed nature of the underlying embeddings}. 
}

% While multiplicative interactions have been utilized in specific contexts within deep learning architectures
While deep learning models have previously utilized multiplicative interactions for specific purposes,  %in specific contexts 
such as fusing information from multiple inputs or mediating 
context-dependent computations (e.g., in FiLM layers, gating mechanisms, and Hypernetworks), these applications have been relatively limited in scope \citep{jayakumarMultiplicativeInteractionsWhere2020}. 
% Our work significantly expands upon this foundation by 
Our work expands the role of multiplicative interactions, demonstrating their effectiveness as the primary computational primitive for modeling abstract relationships between multiple input elements.

% This {\cyan dynamic perspective} contrasts with traditional vector embeddings (e.g., word embeddings), which primarily capture the static semantic meaning of symbols and require additional mechanisms (e.g., self attention) % layers or MLPs) 
% While transformer models can implicitly capture some aspects of dynamic semantics through attention mechanisms, our approach more directly models symbolic interactions through matrix multiplication, potentially offering advantages in interpretability, compositionality, and efficiency.
% % 
% Transformers' dynamic semantics is limited by the static nature of word embeddings, while our approach intrinsically encodes the dynamic nature of meaning into the representation itself.

% By representing symbols as operators, our matrix embedding approach has the potential to offer more direct interaction modeling, improved compositionality, increased interpretability, and greater computational efficiency.
 % compared to traditional vector embeddings.
% In contrast, our matrix embedding approach models each symbol as an operator that directly influences other symbols through matrix multiplication.

% A key challenge in SOC %, unaddressed in prior work, 
% has been the absence of a well-defined complexity metric for symbolic operations, analogous to rank in matrix completion.  Our work addresses this by introducing of a novel regularizer that implicitly defines such a metric, {\green prioritizing} solutions that exhibit general group structures. 

Another core contribution is our novel regularizer, which unlocks the full potential of matrix embeddings. By implicitly promoting unitary representations, this regularizer instills a powerful inductive bias towards discovering general group structures in data. 
% This bias proves critical for inferring symbolic relationships from limited observations, analogous to the low-rank bias in matrix completion.
{\blue This bias, akin to the low-rank bias in matrix completion, provides a novel way to quantify the simplicity of symbolic operations,
% and proves critical for 
and offers an effective solution
for inferring symbolic relationships from limited observations.
}

\paragraph{Symmetry Discovery}

The bias towards general groups % structures %revealed by our model 
offers a promising new paradigm for leveraging %leveraging
symmetries in deep learning. Current approaches often rely on manually designing architectures tailored to specific symmetry groups, such as equivariant networks \citep{bronsteinGeometricDeepLearning2021}. In contrast, our findings suggest a {\green universal} %broader} 
inductive bias towards the fundamental algebraic structures of groups, {\cyan not specific symmetries,} potentially enabling the discovery of symmetries across diverse domains without the need for bespoke architectures.
Notably, the learned representations in HyperCube correspond to the generalized Fourier basis, which mediates %underpins 
the group convolution operation central to equivariant neural networks (see Appendix~\ref{Appendix:fourier_transform_HyperCube}). This connection positions HyperCube as a potential framework for automatically constructing symmetry-aware architectures directly from data.

% In contrast, our findings suggest a universal bias towards the general algebraic structures of groups, not specific symmetries, 
% % presenting a powerful new approach for symmetry discovery across diverse domains.

% our results suggest the possibility of a unified architecture that leverages a universal bias towards groups to automatically discover all types of symmetries directly from data.
% establishing HyperCube as a key building block %for {\red learnable symmetry-aware equivariant networks.
% for automatically discovering the core structure of symmetry-aware equivariant neural networks.

% Since all symmetries form finite or compact groups, this universal bias opens the possibility of a unified deep learning architecture capable of automatically discovering %applicable across 
% Identifying these symmetries, which often reflect fundamental rules or governing principles in nature, can significantly enhance model performance and generalization, and even catalyze new scientific discoveries.

% {\red As symmetries often reflect fundamental {\cyan rules or governing} principles in nature, their identification can significantly enhance model performance and generalization, and even lead to new scientific discoveries.}

\paragraph{Limitation}
While our method exhibits strong sample efficiency, the use of tensor factors can incur substantial memory and computational loads, scaling as $O(n^3)$.
Potential mitigations include exploiting sparsity in the factors (e.g., block-diagonalization) or
utilizing faithful representations of smaller dimensions
(${d \times d}$, $d \ll n$). Even though such representations do not strictly adhere to eq~\eqref{eq:trace_of_regular_representation}, they can still yield correct predictions for the Cayley tables, since they satisfy eq~\eqref{eq:product_tensor} and the maximum value of their characters %({\it i.e.} traces) 
is achieved by the identity element: {\it i.e.}  $\text{argmax}~ \trace[\varrho(g)] = e$.  % when trained with cross-entropy loss instead of squared-error loss.

\paragraph{Open Questions}
This work leaves several open questions for future studies, such as deriving exact generalization bounds for SOC. Additionally, proving Observation~\ref{observation:matrix_slice_unitarity_always} and Conjecture~\ref{conjecture:group_rep_is_minimizer} on the optimality of unitary representations remains an open challenge. Furthermore, scaling the method to problems involving multiple symbols beyond binary operations is an important direction for future research.

In summary, our work establishes group theory as a powerful guiding principle for discovering abstract algebraic structures in deep learning, and showcases matrix representations as a compelling alternative to traditional vector embeddings for modeling symbolic relationships.

% Such matrix-based approach intrinsically encodes the dynamic, interactive nature of meaning into the representation itself, potentially offering advantages in interpretability, compositionality, and efficiency compared to vector-embeddings. % by directly modeling how symbols interact and modify each other's meaning.

% Our novel regularizer, which instills an implicit bias towards learning unitary representations in HyperCube, represents another significant contribution. This regularizer effectively defines a complexity metric that favors general group structures, providing a powerful inductive bias for inferring symbolic relationships from limited observations.
% This metric provides a novel way to quantify the simplicity of symbolic operations, and constitutes an effective way/bias to infer symbolic relationship from limited observations.  

\bibliography{
referenceHuh,
DeepLearningTheory, 
SymmetryLearning,
Grokking,
DeepWeightDecay,
TensorNetwork,
NLP,
MultiplicativeInteractions
} 

\begin{thebibliography}{}

\bibitem[Arora et~al., 2019]{aroraImplicitRegularizationDeep2019a}
Arora, S., Cohen, N., Hu, W., and Luo, Y. (2019).
\newblock Implicit {Regularization} in {Deep} {Matrix} {Factorization}.
\newblock arXiv:1905.13655 [cs, stat].

\bibitem[Bronstein et~al., 2021]{bronsteinGeometricDeepLearning2021}
Bronstein, M.~M., Bruna, J., Cohen, T., and Veličković, P. (2021).
\newblock Geometric {Deep} {Learning}: {Grids}, {Groups}, {Graphs},
  {Geodesics}, and {Gauges}.
\newblock arXiv:2104.13478 [cs, stat].

\bibitem[Burer and Monteiro, 2003]{burerNonlinearProgrammingAlgorithm2003}
Burer, S. and Monteiro, R.~D. (2003).
\newblock A nonlinear programming algorithm for solving semidefinite programs
  via low-rank factorization.
\newblock {\em Mathematical Programming}, 95(2):329--357.

\bibitem[Candes and Tao, 2010]{candesPowerConvexRelaxation2010}
Candes, E.~J. and Tao, T. (2010).
\newblock The {Power} of {Convex} {Relaxation}: {Near}-{Optimal} {Matrix}
  {Completion}.
\newblock {\em IEEE Transactions on Information Theory}, 56(5):2053--2080.

\bibitem[Candès and Recht, 2009]{candesExactMatrixCompletion2009}
Candès, E.~J. and Recht, B. (2009).
\newblock Exact {Matrix} {Completion} via {Convex} {Optimization}.
\newblock {\em Foundations of Computational Mathematics}, 9(6):717--772.

\bibitem[Chughtai et~al., 2023]{chughtaiNeuralNetworksLearn2023}
Chughtai, B., Chan, L., and Nanda, N. (2023).
\newblock Neural {{Networks Learn Representation Theory}}: {{Reverse
  Engineering}} how {{Networks Perform Group Operations}}.
\newblock In {\em {{ICLR}} 2023 {{Workshop}} on {{Physics}} for {{Machine
  Learning}}}.

\bibitem[Fazel et~al., 2001]{fazelRankMinimizationHeuristic2001}
Fazel, M., Hindi, H., and Boyd, S. (2001).
\newblock A rank minimization heuristic with application to minimum order
  system approximation.
\newblock In {\em Proceedings of the 2001 {American} {Control} {Conference}.
  ({Cat}. {No}.{01CH37148})}, volume~6, pages 4734--4739 vol.6.
\newblock ISSN: 0743-1619.

\bibitem[Gunasekar et~al., 2017]{gunasekarImplicitRegularizationMatrix2017}
Gunasekar, S., Woodworth, B.~E., Bhojanapalli, S., Neyshabur, B., and Srebro,
  N. (2017).
\newblock Implicit {Regularization} in {Matrix} {Factorization}.
\newblock In {\em Advances in {Neural} {Information} {Processing} {Systems}},
  volume~30. Curran Associates, Inc.

\bibitem[Heim, 1982]{heimSemanticsDefiniteIndefinite1982a}
Heim, I.~R. (1982).
\newblock The {Semantics} of {Deﬁnite} and {Indeﬁnite} {Noun} {Phrases}.
\newblock {\em Doctoral Dissertations Available from Proquest}, pages 1--426.

\bibitem[Jayakumar et~al., 2020]{jayakumarMultiplicativeInteractionsWhere2020}
Jayakumar, S.~M., Czarnecki, W.~M., Menick, J., Schwarz, J., Rae, J., Osindero,
  S., Teh, Y.~W., Harley, T., and Pascanu, R. (2020).
\newblock Multiplicative {Interactions} and {Where} to {Find} {Them}.

\bibitem[Kamp, 1981]{kampTheoryTruthSemantic1981}
Kamp, H. (1981).
\newblock A {Theory} of {Truth} and {Semantic} {Representation}.
\newblock In {\em Meaning and the {Dynamics} of {Interpretation}}, pages
  329--369. Brill.
\newblock Section: Meaning and the Dynamics of Interpretation.

\bibitem[Liu et~al., 2022]{liuUnderstandingGrokkingEffective2022a}
Liu, Z., Kitouni, O., Nolte, N., Michaud, E.~J., Tegmark, M., and Williams, M.
  (2022).
\newblock Towards {{Understanding Grokking}}: {{An Effective Theory}} of
  {{Representation Learning}}.
\newblock In {\em Advances in {{Neural Information Processing Systems}}}.

\bibitem[Nanda et~al., 2022]{nandaProgressMeasuresGrokking2022}
Nanda, N., Chan, L., Lieberum, T., Smith, J., and Steinhardt, J. (2022).
\newblock Progress measures for grokking via mechanistic interpretability.
\newblock In {\em The {{Eleventh International Conference}} on {{Learning
  Representations}}}.

\bibitem[Power et~al., 2022]{powerGrokkingGeneralizationOverfitting2022}
Power, A., Burda, Y., Edwards, H., Babuschkin, I., and Misra, V. (2022).
\newblock Grokking: {{Generalization Beyond Overfitting}} on {{Small
  Algorithmic Datasets}}.

\bibitem[Recht et~al., 2010]{rechtGuaranteedMinimumRankSolutions2010}
Recht, B., Fazel, M., and Parrilo, P.~A. (2010).
\newblock Guaranteed {Minimum}-{Rank} {Solutions} of {Linear} {Matrix}
  {Equations} via {Nuclear} {Norm} {Minimization}.
\newblock {\em SIAM Review}, 52(3):471--501.

\bibitem[Smolensky, 1990]{smolenskyTensorProductVariable1990}
Smolensky, P. (1990).
\newblock Tensor {Product} {Variable} {Binding} and the {Representation} of
  {Symbolic} {Structures} in {Connectionist} {Systems}.
\newblock {\em Artificial Intelligence}, 46:159--216.

\bibitem[Socher et~al., 2012]{socherSemanticCompositionalityRecursive2012}
Socher, R., Huval, B., Manning, C.~D., and Ng, A.~Y. (2012).
\newblock Semantic {Compositionality} through {Recursive} {Matrix}-{Vector}
  {Spaces}.
\newblock In Tsujii, J., Henderson, J., and Paşca, M., editors, {\em
  Proceedings of the 2012 {Joint} {Conference} on {Empirical} {Methods} in
  {Natural} {Language} {Processing} and {Computational} {Natural} {Language}
  {Learning}}, pages 1201--1211, Jeju Island, Korea. Association for
  Computational Linguistics.

\bibitem[Srebro et~al., 2004]{srebroMaximumMarginMatrixFactorization2004}
Srebro, N., Rennie, J., and Jaakkola, T. (2004).
\newblock Maximum-{Margin} {Matrix} {Factorization}.
\newblock In {\em Advances in {Neural} {Information} {Processing} {Systems}},
  volume~17. MIT Press.

\bibitem[Zhao et~al., 2016]{zhaoTensorRingDecomposition2016}
Zhao, Q., Zhou, G., Xie, S., Zhang, L., and Cichocki, A. (2016).
\newblock Tensor {Ring} {Decomposition}.
\newblock arXiv:1606.05535 [cs].

\end{thebibliography}
\bibliographystyle{apalike}

%%%%%%%%%%%%%%%%%%%%%%%%%%%%%%%%%%%%%%%%%%%%%%%%%%%%%%%%%%%%
\newpage
\appendix

\section{Training Procedure} %\section{Training Details} %Implementation}
% \paragraph{Optimization}
\label{appendix:Training_Procedure}

The factor tensors are initialized with entries randomly drawn from a normal distribution with mean $0$ and standard deviation $1/\sqrt{n}$. Real parameterization ($K=\mathbb{R}$) is used. 
We employ full-batch gradient descent to optimize the regularized loss with learning rate of $0.5$ and momentum of $0.5$. 
For the small scale experiments in Section~\ref{sec:toy_results},
the HyperCube regularizer coefficient is set to $\epsilon=0.1$. 
For the larger scale experiments in Section~\ref{sec:grok_results}, we use $\epsilon=0.05$ for HyperCube and $\epsilon=0.01$ for HyperCube-SE.
Each experiment %are relatively small scale and 
can be quickly run within a few minutes on a single GPU machine. 

\paragraph{$\epsilon$-scheduler}
To overcome the limitations in standard regularized optimization, which often prevents full convergence to the ground truth ($D$), we employ  $\epsilon$-scheduler: 
Once the model demonstrates sufficient convergence ({\it e.g.} the average imbalance falls below a threshold of $10^{-5}$), the scheduler sets the regularization coefficient $\epsilon$ to $0$. This allows the model to fully fit the training data.
The effect of $\epsilon$-scheduler on convergence is discussed in 
Appendix~\ref{appendix:persistence_group_rep}.
Note that $\epsilon$-scheduler only affects the overall scale of model's output vector and does not affect the accuracy of output.

% \paragraph{Anonymized repository}
% % All codes are available at 
% The code for our HyperCube model and experiments is available at 
% \url{https://anonymous.4open.science/r/DeepTensorFactorization4GroupRep-EB92/}.

% % The repository for the baseline Transformer result from 
% The code for the baseline Transformer result from 
% \cite{powerGrokkingGeneralizationOverfitting2022} is available at 
% \url{https://github.com/openai/grok}.

\section{List of Binary Operations}
\label{appendix:binary_operations_list}
Here is the list of binary operations from 
\citet{powerGrokkingGeneralizationOverfitting2022}
that are used in Section~\ref{sec:grok_results}
(with $p = 97$). % and $m=5$).

% SHOWN_IN_GROK = ["addition", "subtraction", "division", "mix1", "quad1", "quad2", "quad3", "cube1", "cube2", "sym_xy", "sym_xyx_inv", "sym_xyx"]

\begin {itemize}
\item (add) $a \circ b = a + b$ (mod $p$) for $0 \leq a, b < p  $. (Group operation)
\item (sub) $a \circ b = a - b$ (mod $p$) for $0 \leq a, b < p  $.
\item (div) $a \circ b = a/b$ (mod $p$) for $0 \leq a < p, 0 < b < p  $.
\item (mix) $a \circ b = [a/b$ (mod $p$) if $b$ is odd, otherwise $a - b$ (mod $p$)] for $0 \leq a, b < p  $.
\item (quad1) $a \circ b = a^2 + b^2$ (mod $p$) for $0 \leq a, b < p  $.
\item (quad2) $a \circ b = a^2 + ab + b^2$ (mod $p$) for $0 \leq a, b < p  $.
\item (quad3) $a \circ b = a^2 + ab + b^2 + a$ (mod $p$) for $0 \leq a, b < p  $.
\item (cube1) $a \circ b = a^3 + ab$ (mod $p$) for $0 \leq a, b < p  $.
\item (cube2) $a \circ b = a^3 + ab^2 + b$ (mod $p$) for $0 \leq a, b < p  $.
\item ($ab$ in $S_5$) $a \circ b = a \cdot b$ for $a, b \in S_5  $.  (Group operation)
\item ($aba\inverse$ in $S_5$) $a \circ b = a \cdot b \cdot a^{-1}$ for $a, b \in S_5  $.
\item ($aba$ in $S_5$) $a \circ b = a \cdot b \cdot a$ for $a, b \in S_5  $.
\end {itemize}

% Note that modular subtraction and division are equivalent up to re-labeling of symbols for a prime modulus $p$. 

\begin{figure}[b]
  \vskip 0.05in
  \begin{center}
    \includegraphics[width=0.55\textwidth]{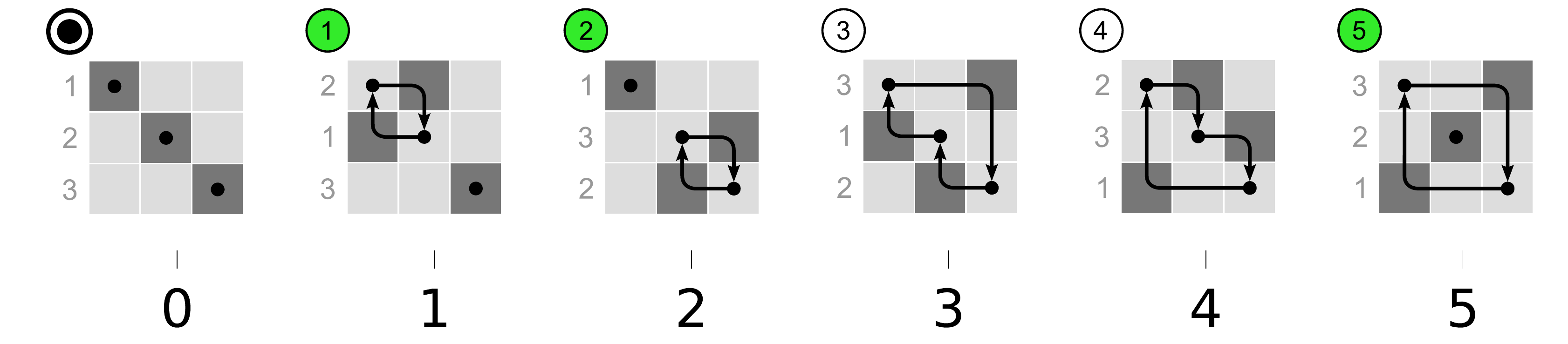}
  \caption{Elements of the symmetric group $S_3$ illustrated as permutations of 3 items. Green color indicates  {\it odd} permutations, and white indicates  {\it even} permutations. Adapted from \url{https://en.wikipedia.org/wiki/Symmetric_group}.}
  \label{fig:S3_elements_illustrated}
  \end{center}
  \vskip -0.2in
\end{figure}

\newpage

\section{Deferred Proofs} % of Lemmas}
\label{Appendix:proofs}

\subsection{Proof of Lemma~\ref{lemma:balanced_condition} on Balanced Condition}
\label{Appendix:balanced}

Here, we derive the balanced condition eq~\eqref{eq:balanced_condition} in detail.  
First, we compute the gradient of the regularized loss 
$  \mathcal L = \mathcal L_o(T)  +\epsilon \mathcal H(A,B,C)  $
eq~\eqref{eq:regularized_loss},
% 
% Gradient terms:
\begin{align}
  \label{eq:grad_A}
  \nabla_{A_a}  \mathcal L & = \frac 1 n ( (\nabla_{T_{abc}}  \mathcal L_o)  \, C_c\trans B_b\trans + 2 \epsilon (A_{a} (B_{b} B_{b}\trans) + (C_{c}\trans C_{c}) A_{a})), \\
  \nabla_{B_b} \mathcal L & = \frac 1 n ( (\nabla_{T_{abc}}  \mathcal L_o)  \, A_a\trans C_c\trans + 2 \epsilon ( B_{b} (C_{c} C_{c}\trans) + (A_{a}\trans A_{a}) B_{b})),  \nonumber \\
  \nabla_{C_c} \mathcal L & = \frac 1 n ( (\nabla_{T_{abc}}  \mathcal L_o)  \, B_b \trans A_a\trans + 2 \epsilon ( C_{c} (A_{a} A_{a}\trans) + (B_{b}\trans B_{b}) C_{c})),  \nonumber
\end{align}
where
% We introduce the following  notation for gradient 
$\nabla_{A_a} \mathcal L  \equiv \partial \mathcal L / \partial{A_a} $,
$\nabla_{B_b}  \mathcal L \equiv \partial \mathcal L / \partial{B_b} $,
$\nabla_{C_c}  \mathcal L \equiv \partial \mathcal L / \partial{C_c} $, and
$\nabla_{T_{abc}}  \mathcal L_o  \equiv \partial \mathcal L_o / \partial{T_{abc}}$.

% Then, it can be easily verified that 
The \emph{imbalances} in 
Lemma~\ref{lemma:balanced_condition}
% Definition~\ref{def:imbalances}
% eq~\eqref{eq:imbalance_terms} 
are defined as the difference of loss gradient:
\begin{align*}
  \xi_I  \equiv 
  \frac {n} {2 \epsilon } (A_a\trans  (\nabla_{a} \mathcal L) - (\nabla_{b} \mathcal L) B_b\trans ) 
  & =  A_a\trans (C_{c}\trans C_{c}) A_a   - B_{b} (C_{c} C_{c}\trans) B_b\trans   \\
  \xi_J  \equiv  
  \frac {n} {2 \epsilon } ( B_b\trans  (\nabla_{b} \mathcal L) - (\nabla_{c} \mathcal L) C_c\trans )
  & =  B_b\trans (A_{a}\trans A_{a}) B_b  - C_{c} (A_{a} A_{a}\trans) C_c\trans   \\
  \xi_{K}  \equiv
  \frac {n} {2 \epsilon } ( C_c\trans (\nabla_{c} \mathcal L) - (\nabla_{a} \mathcal L) A_a\trans )  
  & = C_c\trans (B_{b}\trans B_{b}) C_c  -  A_{a} (B_{b} B_{b}\trans) A_a\trans 
\end{align*} 
At stationary points,   % where gradient vanishes,
{\it i.e. } $\nabla_{A_a} \mathcal L =\nabla_{B_b} \mathcal L =\nabla_{C_c} \mathcal L = 0$,
imbalance terms  vanish to zero, yielding the balanced condition
$\xi_I= \xi_J= \xi_K= 0,$
which proves Lemma~\ref{lemma:balanced_condition}.
Note that imbalance terms are defined to cancel out the $\nabla_{T_{abc}}  \mathcal L_o$ terms.
Therefore, the balanced condition is independent of the loss function 
$\mathcal L_o$.

\subsection{Proof of Lemma~\ref{lemma:Frobenius_fixed_regularizer}}
\label{appendix:Frobenius_fixed_regularizer}
\begin{proof}
    The constraint on Frobenius norm 
    % eq~\eqref{eq:L2_regularizer} 
    can be integrated with the regularizer
    into an augmented loss via the Lagrange multiplier $\lambda$ 
    \begin{align}
      \label{eq:augmented_loss}
      & \mathcal H + \lambda (\mathcal F - {constant}),
    \end{align}
  where %$\mathcal F$ 
  $   \mathcal F \equiv \frac 1 {n}      \trace \left [      A_{a}\trans A_{a}     +  B_{b}\trans B_{b}     +  C_{c}\trans C_{c}    \right ]   $ is the Frobenius norm .

    The gradient of eq~\eqref{eq:augmented_loss} with respect to $A_a$ is
    proportional to
    \begin{align}
      \label{eq:augmented_loss_grad}
      \nabla_{A_a} (\mathcal H + \lambda \mathcal F) \propto
     A_a (B_b B_b\trans) +  (C_c \trans C_c) A_a + \lambda A_a .
  \end{align}
    In the case of C-unitary factors $B$ and $C$,   all terms in eq~\eqref{eq:augmented_loss_grad}      become aligned to $A_a$, {\it i.e.}
    \begin{align}
      \nabla_{A_a} (\mathcal H + \lambda \mathcal F) \propto
      % \alpha_{B}^2 A_a  +  \alpha_{C}^2 A_a + \lambda A_a .
      (\alpha_{B}^2   +  \alpha_{C}^2 + \lambda) A_a .
      % (\sum_b \alpha_{B_b}^2   +  \sum_c \alpha_{C_c}^2 + \lambda) A_a .
    \end{align}    
    and thus an appropriate value for the Lagrange multiplier $\lambda$ can be found to vanish the gradient, which confirms stationarity. 
    This result also applies to gradient with respect to $B_b$ and $C_c$     by the symmetry of parameterization.
  \end{proof}

% \section{Constrained Optimum of Regularizer}

% $$  \mathcal L = \Lambda_{abc}(T_{abc}-D_{abc})  +  \mathcal H(A,B,C) . $$

% $$\nabla_{A_a}  \mathcal L = \frac 1 n ( \Lambda_{abc}  \, C_c\trans B_b\trans + 2  (A_{a} (B_{b} B_{b}\trans) + (C_{c}\trans C_{c}) A_{a} )) $$

\subsection{Persistence of Group Representation}
\label{appendix:persistence_group_rep}

The following lemma demonstrates a key property of our model's convergence behavior: once a group representation is learned, the solution remains within this representational form throughout optimization.

\begin{lemma}
  \label{lemma:persistence_group_rep}
  Let $D$ represent a group operation table. 
  Once gradient descent of the regularized loss eq~\eqref{eq:regularized_loss} converges to a group representation (including scalar multiples),  {\it i.e.}
  \begin{align}
    \label{eq:multiples_of_reg_rep}
    A_{a}  = \alpha_{A_a} \varrho(a), \,
    B_{b} = \alpha_{B_b} \varrho(b), \,
    C_{c} = \alpha_{C_c} \varrho(c)\trans,
  \end{align}
  the solution remains within this representation form.
\end{lemma}

\begin{proof}
With the squared loss eq~\eqref{eq:loss_square}, 
the gradient with respect to $A_a$ eq~\eqref{eq:grad_A} becomes
  \begin{align}
    \label{eq:grad_A_square_loss}
    \nabla_{A_a} \mathcal L & = 
    \frac{1}{n}
    ( \Delta_{abc} M_{abc} C_c\trans B_b\trans + 
    \epsilon (A_a (B_b B_b\trans) +  (C_c \trans C_c) A_a ))  
  \end{align}
where $\Delta \equiv T - D$ is the constraint error, 
and $M$ is the mask indicating observed entries in the train set. 

Substituting the group representation form eq~\eqref{eq:multiples_of_reg_rep} into eq~\eqref{eq:grad_A_square_loss}, we get:
\begin{align}
  \label{eq:grad_A_2nd}
  \frac{1}{n} \epsilon (A_a (B_b B_b\trans) +  (C_c \trans C_c) A_a )  = 2 \epsilon \alpha_{A_a} \alpha^2 \varrho (a),
\end{align}
for the last two terms, %the regularization terms  
where $\alpha^2 = \sum_b \alpha_{B_b}^2 /n =  \sum_c \alpha_{C_c}^2/n$.

Since the product tensor is 
\begin{align*}
  T_{abc} 
  = \frac{1}{n}  \trace [ A_a B_b C_c]  
  = \frac{1}{n} {\alpha_{A_a} \alpha_{B_b} \alpha_{C_c}} \trace [ \varrho(a)\varrho(b)\varrho(c)\trans ] 
  = \alpha_{A_a} \alpha_{B_b} \alpha_{C_c} D_{abc},
\end{align*}
and $D_{abc} = \delta_{a \circ b , c} = \delta_{a , c \circ b\inverse}$
($\delta$ is the Kronecker delta function),
the first term in eq~\eqref{eq:grad_A_square_loss} becomes %reduces to 
\begin{align}
  \frac{1}{n} \sum_{b,c} \Delta_{abc} M_{abc} C_c\trans B_b\trans 
    & = 
    \frac{1}{n} \sum_{b,c} \delta_{a \circ b , c}  M_{abc} (\alpha_{A_a} \alpha_{B_b} \alpha_{C_c} - 1)\alpha_{B_b} \alpha_{C_c}    \varrho (c \circ b\inverse)   \nonumber \\
    \label{eq:grad_A_1st_term}
    & =   \frac{1}{n}  \sum_{b}  M_{ab(a\circ b)}   (\alpha_{A_a} \alpha_{B_b} \alpha_{C_{a \circ b}} - 1)\alpha_{B_b} \alpha_{C_{a \circ b}} \varrho (a)   .
\end{align} 
% where $c$ is replaced with $a \circ b$ for the last equality. 
% 
Note that both  eq~\eqref{eq:grad_A_1st_term} and eq~\eqref{eq:grad_A_2nd} are proportional to $\varrho(a)$.
Consequently, we have 
$\nabla_{A_a} \mathcal L \propto \varrho(a).$  %  \propto A_a
Similar results for other factors can also be derived:
$\nabla_{B_b} \mathcal L \propto \varrho(b) $,
and  $\nabla_{C_c} \mathcal L \propto \varrho(c)\trans $.
This implies that gradient descent preserves the form of the group representation (eq~\eqref{eq:multiples_of_reg_rep}), only updating the coefficients  $\alpha_{A_a} , \alpha_{B_b},\alpha_{C_c}$.
\end{proof}

\paragraph{Effect of  $\epsilon$-Scheduler}

Lemma~\ref{lemma:persistence_group_rep} holds true
even when $\epsilon$ gets modified by  $\epsilon$-scheduler, which reduces $\epsilon$ to $0$.
In this case,  the coefficients converge to $\alpha_{A_a} =  \alpha_{B_b}  = \alpha_{C_c}=1 $,
resulting in the exact group representation form eq~\eqref {eq:factors_are_group_representations}.

\section{Group Convolution and Fourier Transform}
  \subsection{Fourier transform on groups}
  \label{Appendix:Fourier}

  The Fourier transform of a function     $f:G\to \mathbb {R}$   at a representation  $\varrho :G\to \mathrm {GL}(d_{\varrho}, \mathbb {R} )$ of $G$ is 
  \begin{align}
      \label{eq:Fourier_transform}
      {\hat {f}}(\varrho )=\sum _{g\in G} f(g)\varrho (g).
  \end{align}
  
  For each representation $\varrho$ of  $G$,    $\hat {f}(\varrho)$ is a  $d_{\varrho}\times d_{\varrho}$ matrix, where   $d_{\varrho}$ is the degree of $\varrho$. 
  
  \subsection{Dual group}
  \label{Appendix:dual_group}
  
  Let $\hat{G}$ be a complete set indexing the irreducible representations of $G$ up to isomorphism, called the \emph{dual group}, thus for each $\xi$ we have an irreducible representation $\varrho_{\xi}: G \to U(V_{\xi})$, and every irreducible representation is isomorphic to exactly one   $\varrho_{\xi}$. 
  
  \subsection{Inverse Fourier transform} % on Dual group}
  The inverse Fourier transform at an element $g$ of $G$ is given by 
  \begin{align}
      f(g)={\frac {1}{|G|}}\sum _{\xi \in \hat{G}}d_{\varrho_{\xi}}\trace\left[\varrho_{\xi}(g^{-1}){\hat {f}}(\varrho_{\xi})\right].
  \end{align}
  where the summation goes over  the complete set of irreps in $\hat{G}$.

  \subsection{Group Convolution}
  
  The  convolution of two functions  over a finite group   $f,g:G\to \mathbb {R}$ is defined as 
  \begin{align}
      \label{eq:group_convolution}
       (f \ast h)(c) \equiv \sum_{b\in G} f \left(c \circ b\inverse \right) h(b) 
  \end{align}

  \subsection{Fourier Transform of Group Convolution}
  \label{Appendix:model_group_conv}
  
  Fourier transform of a convolution at any representation $\varrho$ of $G$ is given by the matrix multiplication
  \begin{align}
      \label{eq:group_convolution_in_fourier}
      {\widehat {f\ast h}}(\varrho )={\hat {f}}(\varrho ){\hat {h}}(\varrho ).
  \end{align}
  
  In other words, in Fourier representation,   the group convolution is simply implemented by the matrix multiplication.
  
  \begin{proof}
  \begin{align}
      \label{eq:group_convolution_model}
      {\widehat {f\ast h}}(\varrho )
      & \equiv \sum_c \varrho(c) \sum_b f(c \circ b\inverse) h(b) \\
      & = \sum_c \varrho(c) \sum_{a,b} f(a) h(b) \delta_{(a, c \circ b\inverse)} \\
      & =  \sum_{a,b} f(a) h(b) \sum_c \varrho(c) \delta_{(a \circ b, c)} \\
      & =  \sum_{a,b} f(a) h(b)  \varrho(a \circ b)  \\
      & =  \sum_a f(a)\varrho(a) \sum_b  h(b)  \varrho(b)  \\
      & ={\hat {f}}(\varrho ){\hat {h}}(\varrho ).
  \end{align}
  where $\delta$ is the Kronecker delta function,
  and the equivalence between 
  $a = c \circ b\inverse$ and $a \circ b = c $
  is used between the second and the third equality.
  \end{proof}

\section{Group Convolution and Fourier Transform in HyperCube}
\label{Appendix:fourier_transform_HyperCube}

HyperCube shares a close connection with group convolution and Fourier transform.
On finite groups, the Fourier transform generalizes classical Fourier analysis to functions defined on the group:  $f: G \to \mathbb {R}$. Instead of decomposing by frequency, it uses the group's irreducible representations $\{\varrho_\xi \}$, where $\xi$ indexes the irreps (See Appendix~\ref{Appendix:dual_group}). A function's Fourier component at  $\xi$ is defined as:
\begin{align}
    \label{eq:Fourier_component}
    \hat {f}_\xi  \equiv \sum _{g\in G} f(g) \varrho_\xi (g).
\end{align}

\paragraph{Fourier Transform in HyperCube} 

The Fourier transform perspective offers a new way to understand how HyperCube with a group representation eq~\eqref{eq:factors_are_group_representations} processes general input vectors.
Consider a vector $f$ representing a function, {\it i.e.}, $f_g = f(g)$. Contracting $f$ with a model factor $A$ (or $B$) yields:
\begin{align}
    \label{eq:Fourier_transform_by_A}
    \hat{f}  \equiv   
    f_g {A}_{g}  = \sum_{g \in G}  f(g) \varrho(g) ,
\end{align}
which calculates the Fourier transform of $f$  using the regular representation $\varrho$. As $\varrho$ contains all irreps of the group, $\hat{f}$ holds the complete set of Fourier components. Conversely, contracting $\hat{f}$  with $\varrho\trans$ ({\it i.e.} factor $C$) performs the \emph{inverse Fourier transform}:
\begin{align}
    \label{eq:inverse_Fourier_transform_by_A}
    \frac 1 n \trace [\hat{f} {C}_{g}] 
    = \frac 1 n \sum_{g' \in G} f_{g'} \trace [\varrho(g') \varrho(g) \trans]    = f_{g},
\end{align}
where eq~\eqref{eq:trace_of_regular_representation} is used. 
This reveals that the factor tensors generalize the discrete Fourier transform (DFT) matrix, allowing the model to map signals between the group space and its Fourier (frequency) space representations.

% Understanding the model 
Through the lens of Fourier transform, we can understand how the model eq~\eqref{eq:product_tensor} processes general input vectors ($f$ and $h$): it calculates their Fourier transforms  ($\hat{f}, \hat{h}$), multiplies them in the Fourier domain ($\hat{f} \hat{h}$), and applies the inverse Fourier transform. 
Remarkably, this process is equivalent to performing group convolution ($f \ast h$).
This is because the linearized group operation (Section~\ref{sec:modeling_framework}) naturally entails  group convolution (see Appendix~\ref{appendix:hypercube_group_convolution},\ref{appendix:group_convolution_by_D}). 

This connection reveals a profound discovery: HyperCube's ability to learn symbolic operations is fundamentally the same as learning the core structure of group convolutions. This means HyperCube can automatically discover the essential architecture needed for equivariant networks, without the need to hand-design them. This finding highlights the broad potential of HyperCube's inductive bias, extending its applicability beyond the realm of symbolic operations.

\subsection{Reinterpreting HyperCube's computation} %$\tilde{\mathcal{T}}$}
\label{appendix:hypercube_group_convolution}
  
  HyperCube equipped with group representation eq~\eqref{eq:product_tensor} processes general input vectors $f$ and $h$ as 
  \begin{align}
      \label{eq:model_group_conv}
      f_a h_b T_{abc} 
      &=  \frac{1}{n}  \sum_{a} \sum_{b} f(a) h (b) \trace  \left[  \varrho(a)   \varrho(b)   \varrho(c)\trans  \right] \nonumber \\
      &= \frac{1}{n}  \trace  \left[  \left(\sum_{a} \varrho(a) f(a) \right) \left(\sum_{b} \varrho(b) h (b) \right)  \varrho(c)\trans  \right] \nonumber \\
      &= \frac{1}{n}  \trace [  (\hat{f} \hat{h}) \varrho(c)\trans ] %\nonumber \\
      = \frac{1}{n}  \trace [  {\widehat {f\ast h}} \, \varrho(c)\trans ] \nonumber \\
      & = (f \ast h)_c.
  \end{align}
  Therefore, the model calculates the Fourier transform of the inputs ($\hat{f}$ and $\hat{h}$), multiplies them in the Fourier domain ($\hat{f} \hat{h}$), and applies the inverse Fourier transform, which is equivalent to the group convolution, as shown in Appendix~\ref{Appendix:model_group_conv}.
  
  \subsection{Group Convolution by $D$}
  \label{appendix:group_convolution_by_D}
 
  Here we show that the linearized group operation $\tilde{D}$ in  Section~\ref{sec:modeling_framework}
  is equivalent to the group convolution in Appendix~\ref{Appendix:model_group_conv}.
  
  Consider contracting the data tensor $D$ with two functions %vectors,
  $f,h \in G$, as 
  % derived the data tensor $D$ as a vector-space representation of a group operation,   using the one-hot encoded basis vectors. 
%   
  % This tensor mediates the group convolution between two functions %vectors,
  % $f$ and $h$, as 
  % 
  \begin{align}
      \label{eq:group_convolution_by_D}
      f_a h_b  D_{abc}
      = \sum_{ab} f(a) h(b)  \delta_{(a, c \circ b\inverse)}
      = \sum_{b} f(c \circ b\inverse) h(b) 
      \equiv (f \ast h)(c),
  \end{align}
  which computes the group convolution between $f$ and $h$, 
  similar to eq~\eqref{eq:model_group_conv}.
  Here, we used   $D_{abc} = \delta_{(a \circ b, c)} = \delta_{(a, c \circ b\inverse)}$.

\newpage
  
  \section{Supplemental Figures for Section~\ref{sec:toy_results}}

  \begin{figure}[h]
    \vskip 0.05in
    \begin{center}

      \includegraphics[width=0.99\textwidth]{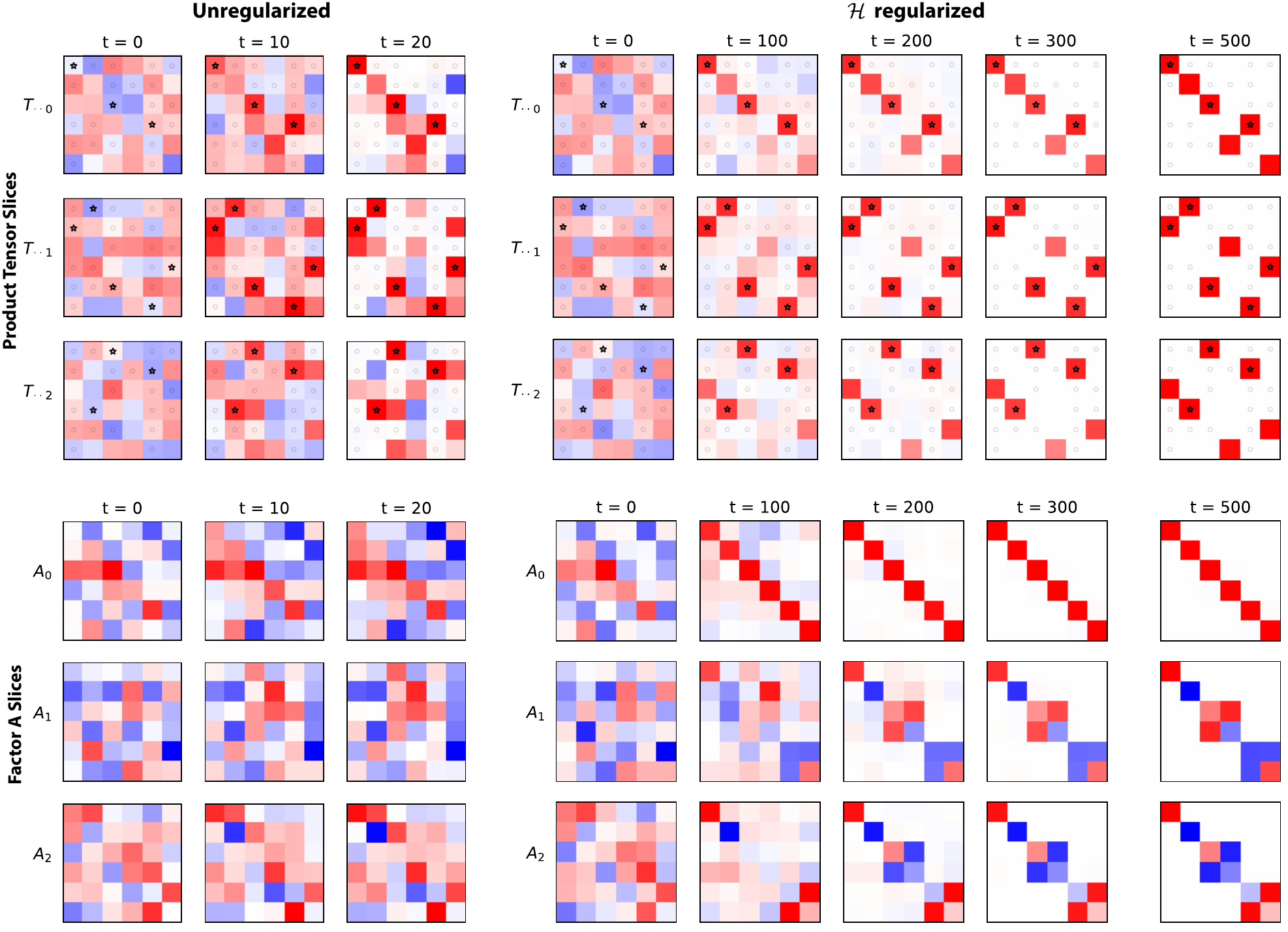}

      \vskip 0.05in

      \caption{
      Visualization of  the %end-to-end 
      model $T$ and factor $A$ during training on the symmetric group $S_3$
      % with (Right) and without (Left) regularization      
      (see Fig~\ref{fig:sym3_training_traj}).
      (Top) Model slices: $T_{\cdot\cdot c}$.
      % The training data $\Omega_\text{train}$ are marked by asterisks (1s) and light gray circles (0s).
      %  (60\% of the total data).
      % 
      The unregularized model 
      quickly converges to a solution with poor generalization. % (Left).
      %  quickly learns the train dataset, but shows little generalization on the test dataset.
      The $\mathcal H$-regularized model %(Right) 
      converges to a generalizing solution around $t=200$. % achieving perfect recovery of the unmarked test data. 
      It accurately recovers $D$ when 
      the regularization diminishes around $t=400$  ($\epsilon \to 0$).
      (Bottom)    Factor $A$ slices: $A_a$. The unregularized model shows minimal changes from random initial values, while $\mathcal H$-regularized model shows significant internal restructuring.
      Shown in the block-diagonalizing  coordinate. See Fig~\ref{fig:basis_change} (Bottom).
      Only the first three slices are shown. %of $T$ and $A$ are shown. % for brevity.
      (color scheme: red=1, white=0, blue=-1.)
      }
    \label{fig:sym3_weight_hist}
    \end{center}
    \vskip -0.5in
  \end{figure}

  \begin{figure}[h]
    % \vskip 0.05in
    \begin{center}
      \includegraphics[width=0.8\textwidth]{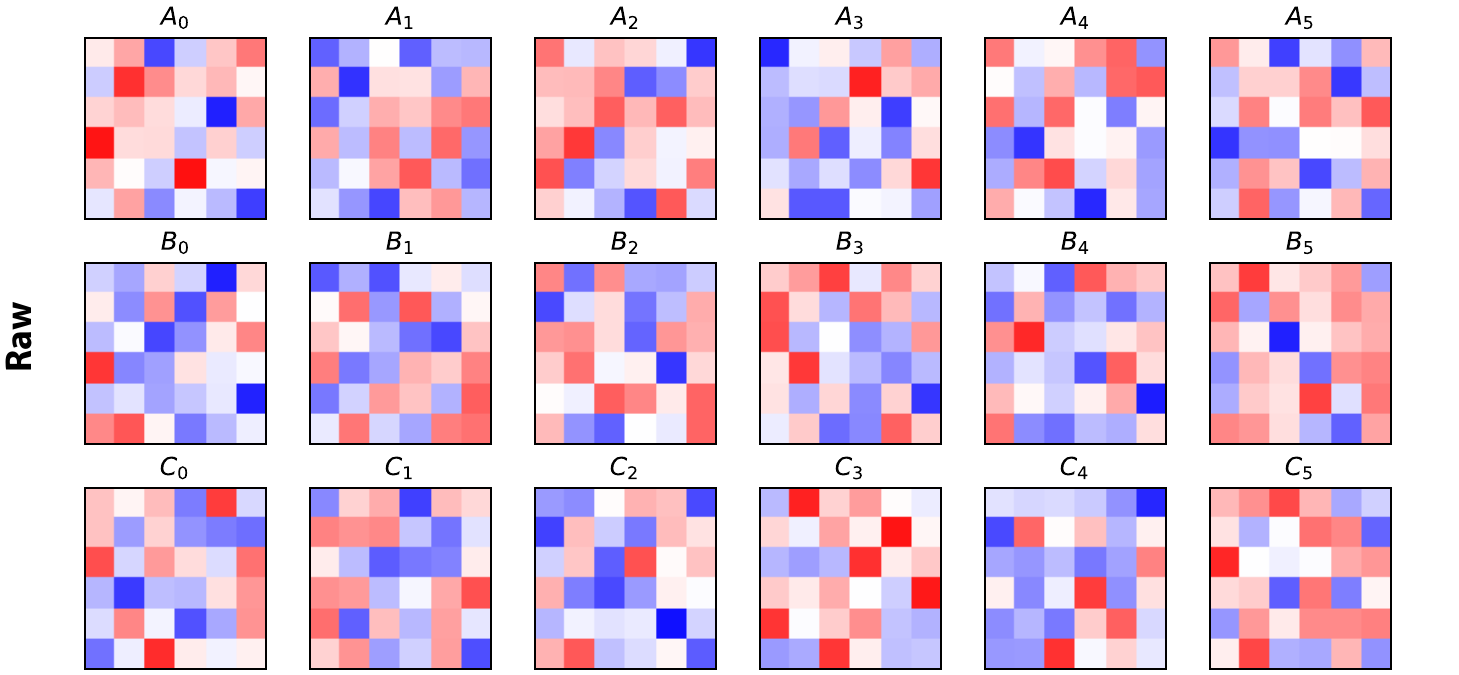}
  
      \vspace*{0.4 cm}
  
      \includegraphics[width=0.8\textwidth]{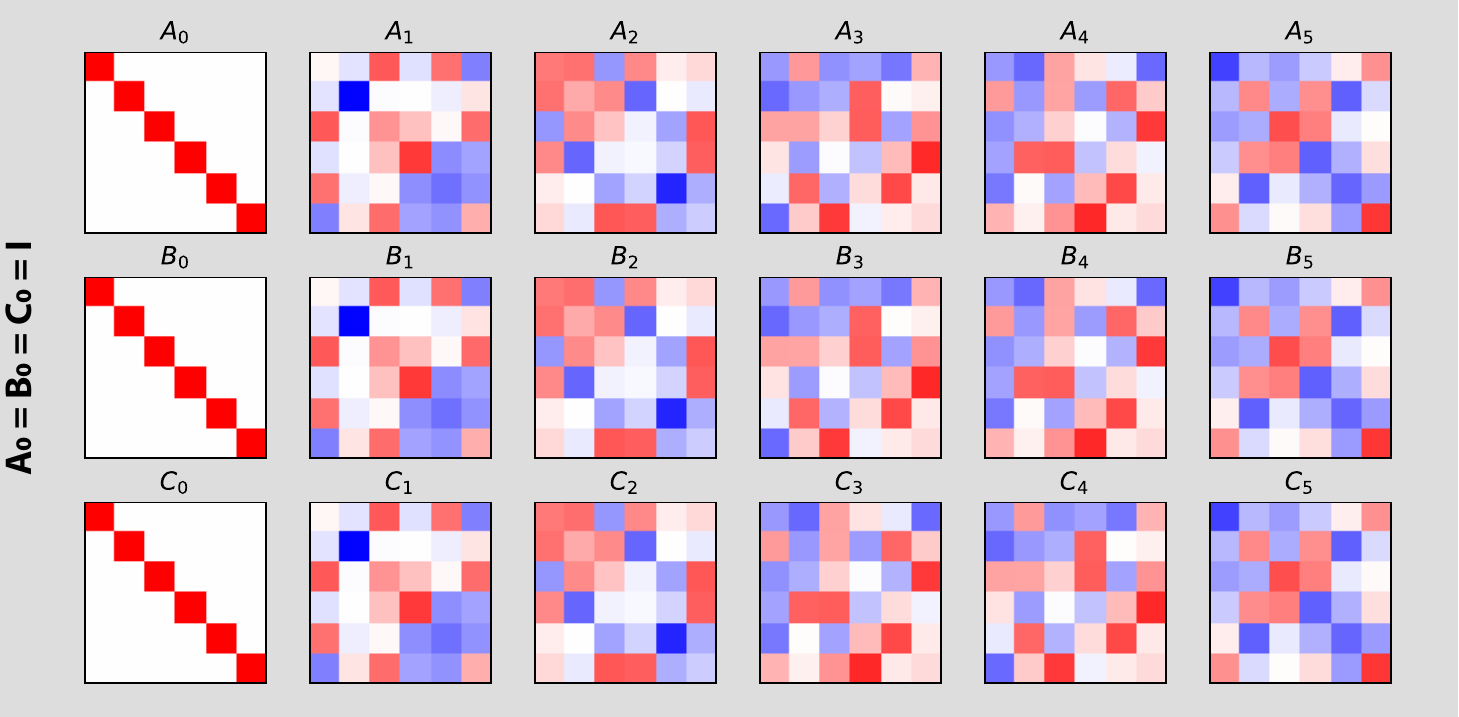}
  
      \vspace*{0.4 cm}
  
      \includegraphics[width=0.8\textwidth]{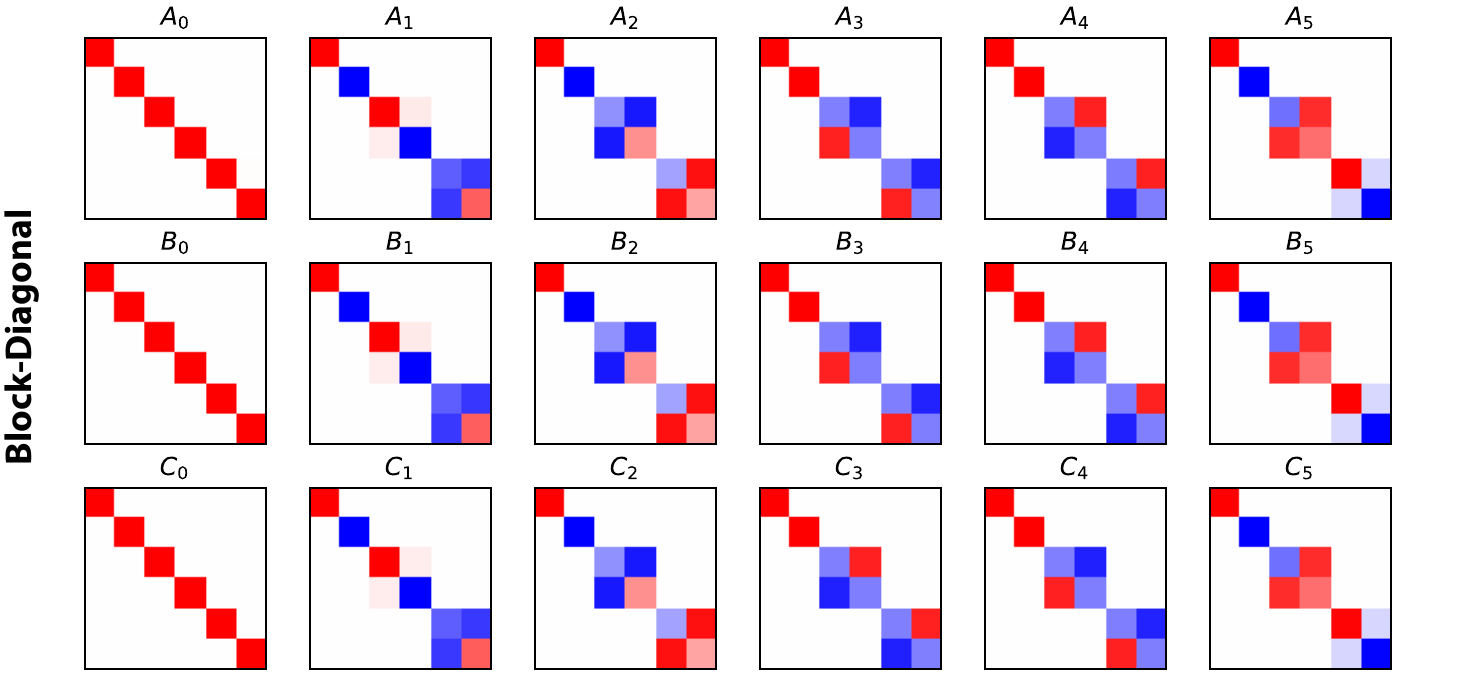}
          \caption{
      Learned factors of the $\mathcal H$ regularized model trained on the $S_3$ group.
       (Top) Raw factor weights shown in their native {coordinate representation}. %Each matrix slice is unitary matrix. 
      (Middle)      Unitary basis change using       $U_I = I$, $U_K = A_0$, $U_J = B_0 \trans$,         which yields $\tilde{A}_0 = \tilde{B}_0 = \tilde{C}_0 = I$.
      (Bottom)       Factors in a block-diagonalizing basis coordinate, revealing their decomposition into direct sum of irreducible representations.
      (color scheme: red=1, white=0, blue=-1.)
    }
    \label{fig:basis_change}
    \end{center}
    % \vskip -0.2in
  \end{figure}

  \begin{figure}[h]
    % \vskip 0.05in
    \begin{center}
      \includegraphics[width=0.65\textwidth]{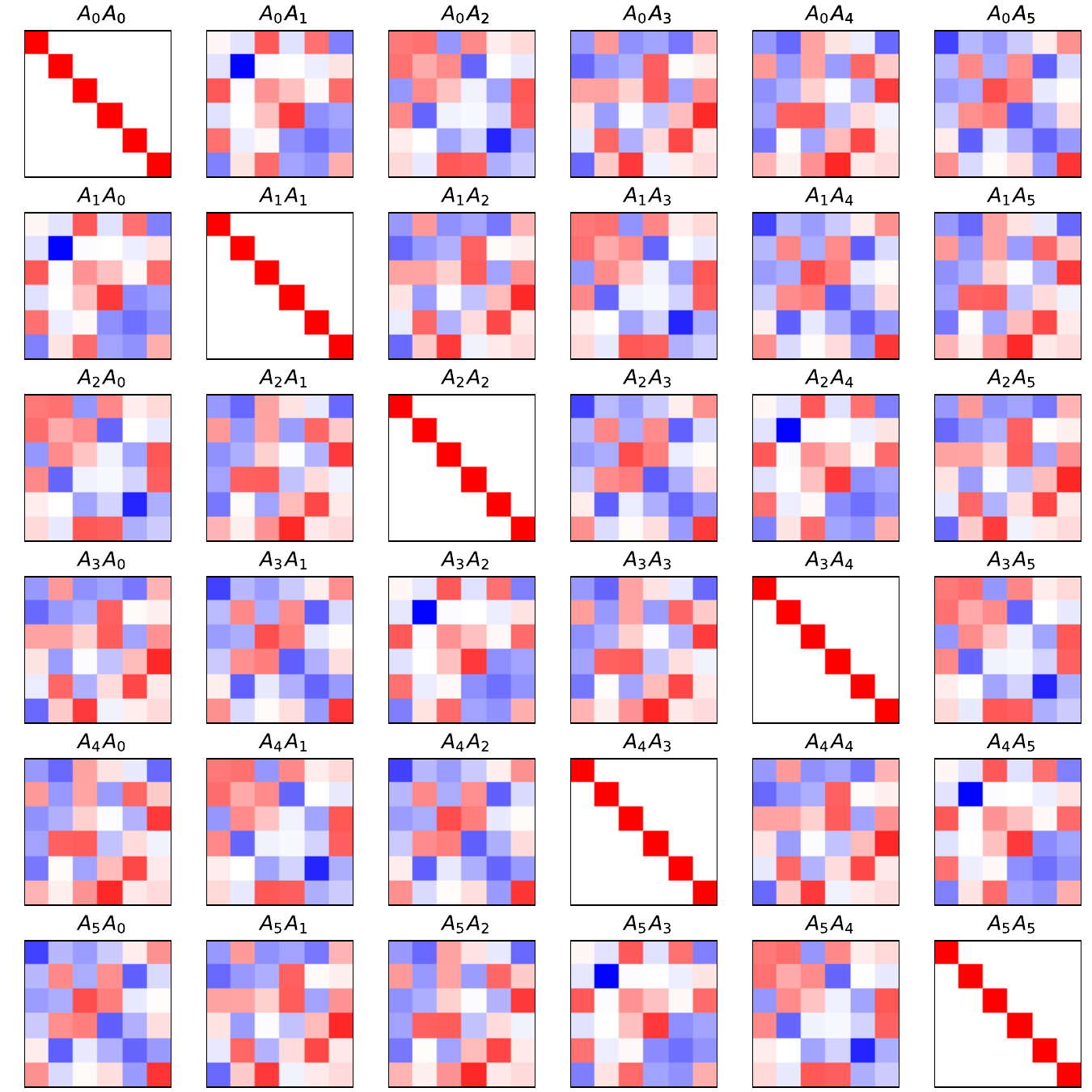}
    \caption{ Multiplication table of matrix slices of factor $A$ from the mid panel of Fig~\ref{fig:basis_change}. 
    Note that this table share the same structure as the Cayley table of the symmetric group $S_3$ in Fig~\ref{fig:cayley_tables}.
    (color scheme: red=1, white=0, blue=-1.)
  }
    \label{fig:AA_multiplication_table}
    \end{center}
    % \vskip -0.2in
  \end{figure}

\clearpage 
  
  \begin{figure}[h]
    % \vskip -0.1in
    \begin{center}
      \includegraphics[width=0.95\textwidth]{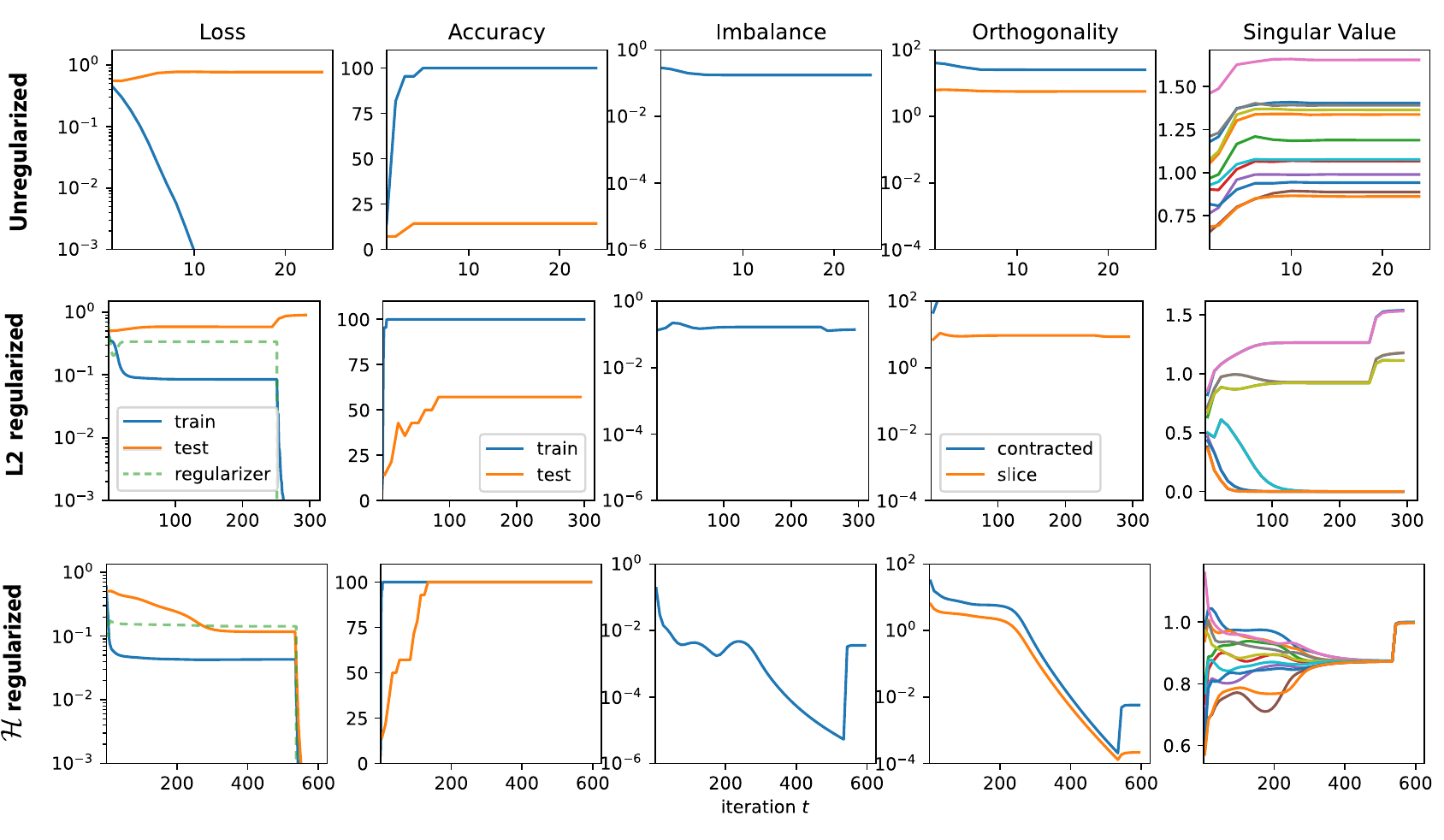}
      \vskip -0.1in
      \caption{
      Optimization trajectories on the modular addition (cyclic group $C_6$)  dataset, with 60\% of the Cayley table used as train dataset (see Fig~\ref{fig:add6_T}). %{fig:sym3_weight_hist}).
      (Top) Unregularized, (Middle) L2-regularized, and (Bottom) $\mathcal H$-regularized training.
      }
    \label{fig:add6_training_traj}
    \end{center}
    \vskip -0.1in
  \end{figure}

  \begin{figure}[h]
    % \vskip -0.1in
    \begin{center}
      \includegraphics[width=0.65\textwidth]{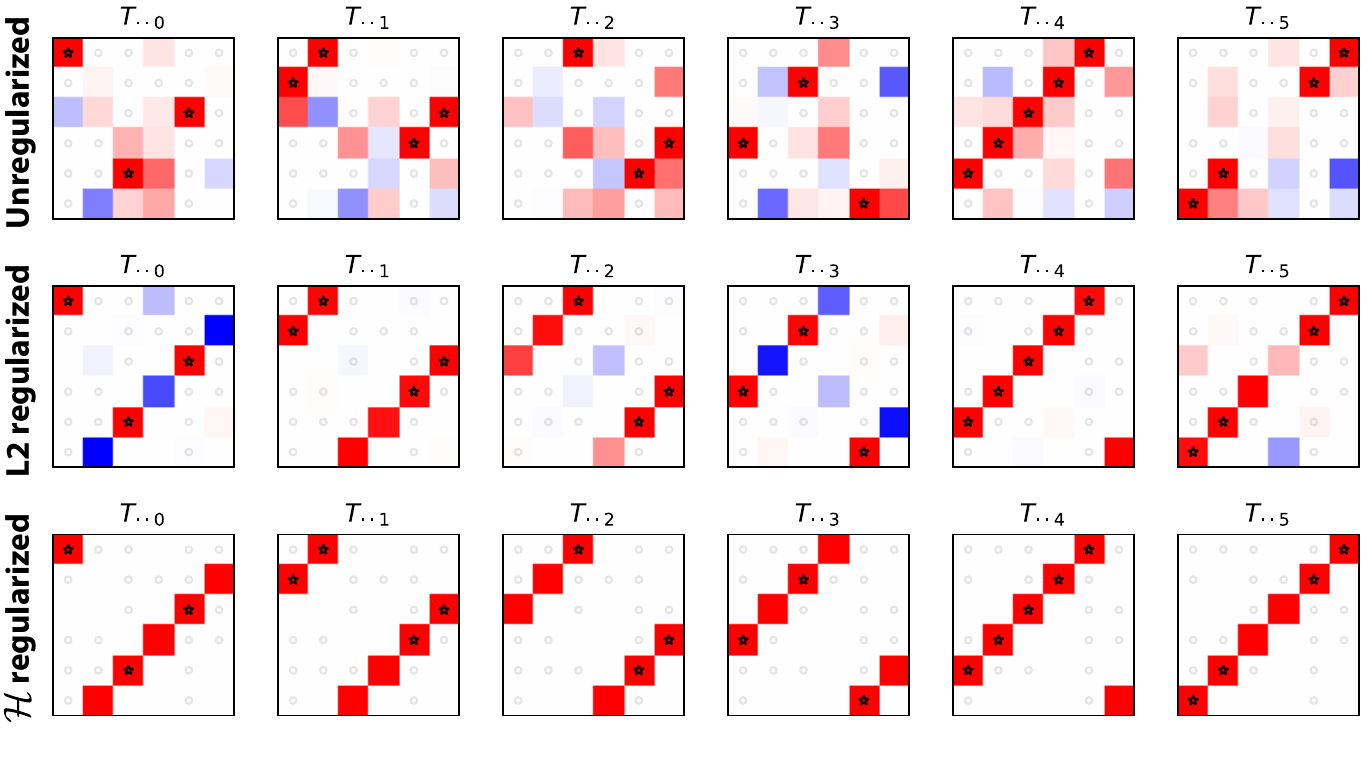}
      % \vskip -0.1in
      \caption{Visualization of product tensors after training on the modular addition (cyclic group $C_6$) under different regularization strategies (see Fig~\ref{fig:add6_training_traj}). 
    The observed training data are marked by asterisks (1s) and {gray} circles (0s).
    Only the $\mathcal H$ regularized model shows perfect recovery of the data tensor $D$. %the unmarked test data. 
    (color scheme: red=1, white=0, blue=-1.)
  }
  \label{fig:add6_T}
    \end{center}
    \vskip -0.4in
  \end{figure}

  \begin{figure}[h]
    \vskip 0.05in
    \begin{center}
      \includegraphics[width=0.72\textwidth]{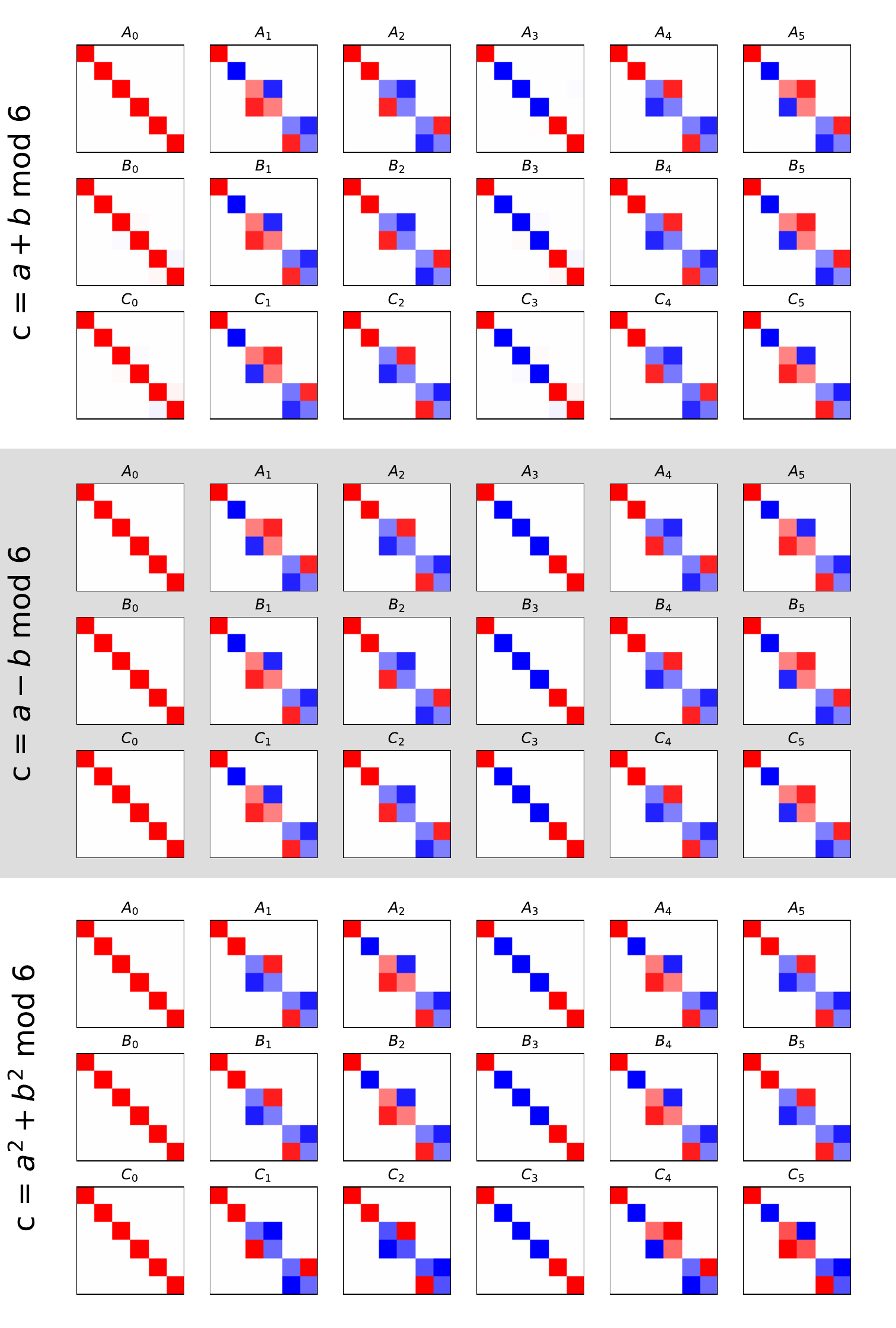}
      
    \caption{
      Visualization of factors       trained on small Cayley tables from Figure~\ref{fig:cayley_tables}. 
      (Top) $c = a+b$ mod $6$,      satisfying      ${A}_g={B}_g={C}_g\trans =  \varrho(g).$
      (Middle) $c = a-b$ mod $6$,       satisfying      ${A}_g\trans={B}_g={C}_g =  \varrho(g).$
      (Bottom) $c = a^2+b^2$ mod $6$,     
      which exhibits the same representation as modular addition
      for elements with unique inverses   (e.g., $g=0, 3$).
      For others,    it learns {\it duplicate} representations reflecting the periodicity of squaring modulo $6$:
      {\it e.g.},  $A_2 = A_4$ and $A_1 = A_5$,      since $2^2 = 4^2$   and $1^2 = 5^2$.      
      (color scheme: red=1, white=0, blue=-1.)
    }
    \label{fig:+-quad}
    \end{center}
    \vskip -0.2in
  \end{figure}

% Optionally include supplemental material (complete proofs, additional experiments and plots) in appendix.
% All such materials \textbf{SHOULD be included in the main submission.}

%%%%%%%%%%%%%%%%%%%%%%%%%%%%%%%%%%%%%%%%%%%%%%%%%%%%%%%%%%%%

% \clearpage
% \input{../Neurips_checklist.tex}

\end{document}